\documentclass[lettersize,journal]{IEEEtran}
\usepackage{amsmath,amsfonts}
\usepackage{algorithm}
\usepackage{algpseudocode}
\usepackage{array}
\usepackage[caption=false,font=normalsize,labelfont=sf,textfont=sf]{subfig}
\usepackage{textcomp}
\usepackage{stfloats}
\usepackage{url}
\usepackage{verbatim}
\usepackage{graphicx}
\usepackage{cite}
\usepackage{multicol}
\usepackage{bm}
\usepackage{amssymb}
\usepackage{booktabs} 

\begin{document}

\title{GLRT-Based Metric Learning for Remote Sensing Object Retrieval}

\author{Linping Zhang, ~\IEEEmembership{Student Member, IEEE,} Yu Liu, ~\IEEEmembership{Member, IEEE,} Xueqian Wang, ~\IEEEmembership{Member, IEEE,} Gang Li, ~\IEEEmembership{Senior Member, IEEE,} and You He 
\thanks{This work was co-supported by National Key R\&D Program of China under Grant 2021YFA0715202, National Natural Science Foundation of China under Grants 62425117, 62101303, 62293540, 62341130, 62471274, Autonomous Research Project of Department of Electronic Engineering at Tsinghua University and the Outstanding Youth Innovation Team Program of University in Shandong Province (2021KJ005). Corresponding author: Yu Liu (liuyu77360132@126.com).

L. Zhang, X. Wang, G. Li, Y. He, and Y. Liu are with the Department of Electronic Engineering, Tsinghua University, Beijing 100084, P.R. China (E-mail: zlp22@mail.tsinghua.edu.cn; wangxueqian@mail.tsinghua.edu.cn; gangli@tsinghua.edu.cn; heyou\_f@126.com; liuyu77360132@126.com).}
}



\maketitle

\begin{abstract}
With the improvement in the quantity and quality of remote sensing images, content-based remote sensing object retrieval (CBRSOR) has become an increasingly important topic. However, existing CBRSOR methods neglect the utilization of global statistical information during both training and test stages, which leads to the overfitting of neural networks to simple sample pairs of samples during training and suboptimal metric performance. Inspired by the Neyman-Pearson theorem, we propose a generalized likelihood ratio test-based metric learning (GLRTML) approach, which can estimate the relative difficulty of sample pairs by incorporating global data distribution information during training and test phases. This guides the network to focus more on difficult samples during the training process, thereby encourages the network to learn more discriminative feature embeddings. In addition, GLRT is a more effective than traditional metric space due to the utilization of global data distribution information. Accurately estimating the distribution of embeddings is critical for GLRTML. However, in real-world applications, there is often a distribution shift between the training and target domains, which diminishes the effectiveness of directly using the distribution estimated on training data. To address this issue, we propose the clustering pseudo-labels-based fast parameter adaptation (CPLFPA) method. CPLFPA efficiently estimates the distribution of embeddings in the target domain by clustering target domain instances and re-estimating the distribution parameters for GLRTML. We reorganize datasets for CBRSOR tasks based on fine-grained ship remote sensing image slices (FGSRSI-23) and military aircraft recognition (MAR20) datasets. Extensive experiments on these datasets demonstrate the effectiveness of our proposed GLRTML and CPLFPA.

\end{abstract}


\section{Introduction}

\begin{figure*}[!t]
\centering
\includegraphics[width=7.2in]{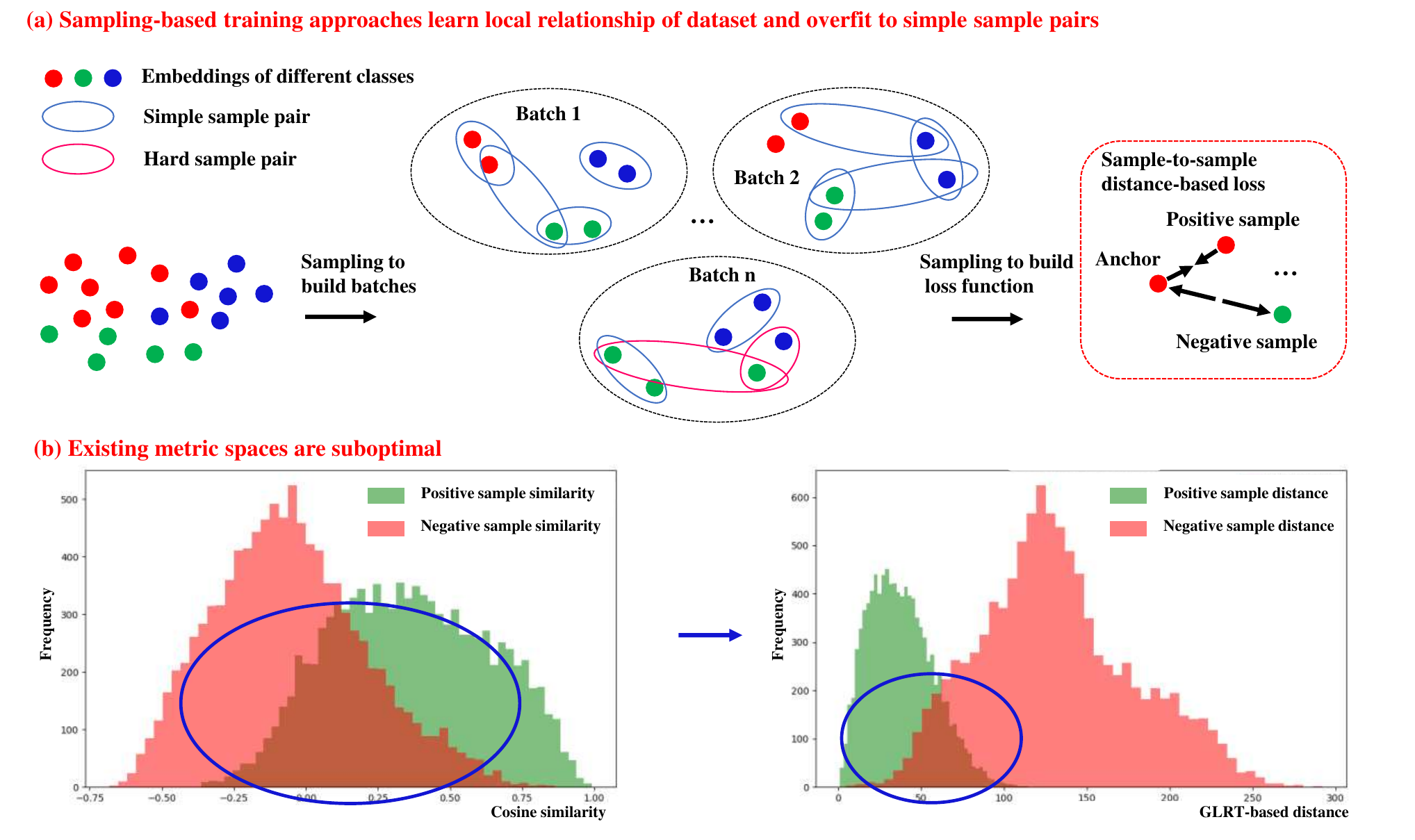}
\caption{Disadvantages of the existing CBRSOR methods. (a) is the visualization of the distance between positive and negative sample pairs on CBRSOR-FGSRSI dataset. The graph on the left of (a) is the result of a cosine similarity-based measure, the higher the similarity of positive samples and the lower the similarity of negative samples, the better. The graph on the right of (a) is the result of our proposed GLRT-based metric learning, where the smaller the distance between positive sample pairs and the larger the distance between negative sample pairs, the better. (b) shows existing methods build loss functions based on sampling data, where simple samples dominate. This phenomenon causes neural networks to only learn the local relationships of the dataset and easily overfit to simple samples. }
\label{fig_1}
\end{figure*}

\IEEEPARstart{W}{ith} the rapid advancement of the remote sensing field, the volume of available remote sensing images has increased significantly. As a result, there is a growing demand for automated methods to retrieve relevant object images from large-scale remote sensing databases based on a given query object slice \cite{ref1}, called content-based remote sensing object retrieval (CBRSOR). While scene-level content-based remote sensing image retrieval (SL-CBRSIR) has seen considerable development in recent years \cite{ref2, ref3, ref4, ref5, ref6, ref7}, it only distinguishes different scenarios, which falls far short of the need to retrieve specific objects. A few works have paid attention to CBRSOR \cite{ref1, ref8, ref9}. Xiong et al. \cite{ref9} develop a cognitive network for fine-grained ship classification and retrieval in remote sensing images. In \cite{ref1}, a fusion Siamese network is proposed for addressing the multi-modality ship remote sensing image retrieval. Guo et al. \cite{ref8} propose a distillation-based hashing transformer to improve the precision of cross-modal ship retrieval.

However, existing CBRSOR methods often fail to effectively utilize global statistical information during both the training and testing phases, leading to overfitting on simple sample pairs during training and suboptimal performance during testing. As shown in Fig. 1(a), due to the memory limitations of the graphics processing unit (GPU), current approaches are restricted to processing data in batches during the training phase, where the loss function is computed over a limited subset of the data. This limitation, combined with the prevalence of simple sample pairs in the dataset, causes the model to overfit to these simple sample pairs, preventing the network from adequately learning discriminative features from more informative and challenging sample pairs. As a result, the network exhibits suboptimal performance on CBRSOR tasks. This highlights the need to incorporate global statistical information during training to emphasize difficult sample pairs and improve the generalization ability of the model. Recent studies have attempted to improve metric learning by incorporating global data relationships during training. For example, Wen et al. \cite{ref21} introduced center loss, which maintains and updates class center embeddings at each iteration. Similarly, Cai et al. \cite{ref22} proposed a class-wise center bank that contrasts class center embeddings with query embeddings using contrastive loss. These center embedding-based methods \cite{ref21,ref22} encourage the neural network to learn global data relationships to some extent by considering the relationship between training data and class centers, rather than focusing on individual data points. However, the ability of these methods to fully capture global data relationships remains limited because the computation of class centers inevitably leads to the loss of intra-class variability. Furthermore, the pairs formed between data points and class centers are often easier and tend to stabilize in the later stages of training, reducing the model's capacity for further optimization and leaving it vulnerable to overfitting. In contrast, difficult sample pairs are critical for continuously refining the network's generalization capabilities.

Second, as depicted in Fig. 1(b), traditional metric spaces, such as cosine similarity, is insufficient for distinguishing between positive and negative sample pairs because they do not account for the global data distribution. In response, several new metric spaces \cite{ref66,ref67} have been proposed to better capture the relationships between embeddings. Manifold-based metric learning \cite{ref66}, for instance, leverages the underlying manifold structure of data to improve the discriminative power of learned embeddings. Despite its potential, this approach struggles with large datasets due to the computational burden associated with estimating manifold similarity through nearest neighbor graphs. In addition, manifold-based metric learning still lacks a comprehensive mechanism to account for global data distribution, which limits its overall effectiveness. Furthermore, mutual information-based metric learning \cite{ref67} attempts to address these issues by maximizing the mutual information between similar samples, thus optimizing the embeddings and reducing neighborhood ambiguity during the training phase. However, this method poses implementation challenges during the testing phase. The lack of statistical information during testing, combined with the inconsistency between the training and testing metric spaces, ultimately leads to suboptimal performance.

The Neyman-Pearson theorem \cite{ref56} proves that when the estimation of hypothesis distributions is accurate, the metric space based on the likelihood ratio test performs the optimal hypothesis test at a given false alarm rate (i.e., identifying unpaired query and gallery images as paired) by fully utilizing global statistical information of data. This theorem motivates our exploration of applying the GLRT to metric learning and the proposal of GLRTML.  In GLRTML, we model the retrieval problem as a hypothesis testing problem, i.e., whether the query image and gallery image are unpaired (null hypothesis, $H_1$), or paired (alternative hypothesis, $H_0$). Naturally, the likelihood ratio between hypothesis $H_1$ and hypothesis $H_0$ can be used as the similarity score, and the CBRSOR task is completed by ranking the similarity scores. The calculation of likelihood ratio requires the estimation of null hypothesis distribution and alternative hypothesis distribution. Therefore, we assume that null hypothesis distribution and alternative hypothesis distribution obey different multivariate distributions, and estimate the distribution parameters through maximum likelihood estimation (MLE). We construct the loss function based on the likelihood ratio, thereby introducing the null hypothesis distribution and alternative hypothesis distribution of the entire training set during the training process. GLRTML naturally estimates the difficulty of each sample pair by evaluating its probability distribution. This allows the network to focus more on difficult samples during training, guiding it to learn more discriminative and robust feature embeddings. By incorporating global statistical information, GLRTML enhances the network’s overall generalization capability. In addition, GLRTML leverages the likelihood ratio test with global data distribution information to construct a more effective metric space during test stage.

Accurately estimating both the null hypothesis and alternative hypothesis distributions is essential for the effectiveness of GLRTML. When the training and test sets share the same distribution, the distribution estimated from the training set can be directly applied to the test set. However, in real-world scenarios, domain discrepancies often arise, where the test set distribution differs from that of the training set. This domain bias can lead to inaccurate estimates from the training data and, consequently, a significant decline in the performance of GLRTML. A widely adopted approach to addressing domain bias is unsupervised domain adaptation (UDA) \cite{ref6, ref12, ref13, ref14, ref15, ref16, ref17, ref18, ref46, ref47, ref48, ref49, ref51, ref52, ref53, ref54}, which adjusts the parameters of the neural network to adapt the model to the target domain. Inspired by pseudo-label-based UDA methods \cite{ref6, ref17, ref49, ref14, ref51, ref52, ref53, ref54}, we propose the clustering pseudo-labels-based fast parameter adaptation (CPLFPA) method to tackle the domain discrepancies. CPLFPA works by first generating embeddings for target domain images using a pre-trained model and clustering these embeddings to assign pseudo-labels. The hypothesis distributions are then re-estimated based on these pseudo-labels. Unlike existing UDA approaches, CPLFPA focuses solely on re-estimating the parameters of the hypothesis distributions, leading to a significant reduction in computational complexity. For example, when using an embedding length of 64 and assuming multivariate Gaussian distributions, only around 4K parameters need to be adjusted. This results in a substantial reduction in computational cost compared to existing UDA methods.

In this article, we propose GLRTML for the task of CBRSOR, and our contributions are summarized as three-fold:

1)	We propose a generalized likelihood ratio test-based metric learning (GLRTML) framework for content-based remote sensing object retrieval (CBRSOR), which incorporates statistical distributions of the entire dataset during both the training and testing stages to construct a more robust metric space. GLRTML estimates the difficulty of sample pairs by leveraging global statistical information, guiding the network to focus on challenging samples. This approach mitigates overfitting and enhances the model's overall generalization performance.

2)	We propose the clustering pseudo-labels-based fast parameter adaptation (CPLFPA) method, which addresses the challenge of re-estimating the distribution of GLRTML in the situation of domain discrepancy with minimal computational effort.

3)	We conduct extensive experiments to validate the effectiveness of our proposed GLRTML and CPLFPA on both CBRSOR and UDA CBRSOR tasks. Our methods achieve state-of-the-art (SOTA) performance, demonstrating their superior capability in handling these challenging tasks.

The rest of this paper is organized as follows. Section II reviews related works. Section III describes the proposed CBRSOR method, including GLRTML and CPLFPA. Experimental results and empirical analysis of the proposed algorithm are reported in Section IV. Section V finally concludes the paper.

\section{Related Work}
In this section, we introduce the most related work from three perspectives: 

\subsection{Content-based Remote Sensing Object Retrieval}
Content-based remote sensing object retrieval (CBRSOR) requires fine-grained discrimination at the object level, such as distinguishing between different types of aircraft and ships. This task is inherently more challenging than scene-level content-based remote sensing image retrieval (SL-CBRSIR) because it requires the retrieval of specific objects with detailed features rather than broad scenes. Due to the success of deep learning models and their relevance to our work, our review focuses primarily on deep learning-based approaches to CBRSOR. Object-level information typically contains rich details and distinct geometric features. However, convolutional neural networks (CNNs), which are widely used in deep learning, may not be ideal for CBRSOR due to their limited ability to model global context and the potential information loss during pooling operations. To address these challenges, Zhao et al. \cite{ref8} propose the use of Transformers to capture global information and design a distillation network to enhance the extraction of fine-grained details from different modalities of data. Xiong et al. \cite{ref9} introduce an interpretable attention-based feature representation module that visualizes areas of interest, and helps the retrieval model to focus on discriminative regions of ships. In another work, Xiong et al. \cite{ref1} present Cog-Net, an interpretable model specifically designed for fine-grained ship classification and retrieval in remote sensing images. Cog-Net integrates cognitive processes into decision making by exploiting fine-grained information, allowing it to perform more accurate retrieval tasks. In addition, the fine-grained ship remote sensing image slices (FGSRSI-23) dataset introduced in \cite{ref1} has been made available to encourage further research in the CBRSOR community. 

Despite these advances, existing methods \cite{ref1,ref8,ref9} overlook the use of global information and rely on sampling to learn relationships between local data during the training phase, primarily due to GPU memory limitations. This reliance introduces performance bottlenecks because simple samples often dominate the dataset, and the relationships learned from these samples tend to be less discriminative, reducing the network's ability to generalize effectively. As a result, the networks overfit to frequently encountered simple samples, leading to degraded retrieval performance. In this article, we present a novel approach that incorporates global statistical information from the data embeddings, allowing the model to evaluate the relative importance of each data point. This strategy prevents the network from overfitting to common simple patterns and improves its overall performance.

\subsection{Metric Learning for Retrieval}
Traditional metric learning methods aim to estimate the magnitude or weight of each component of the feature vector when computing distances, thereby highlighting the most discriminative features. These approaches include unsupervised methods such as principal component analysis (PCA) \cite{ref25}, locality preserving projections (LPP) \cite{ref26}, and multidimensional scaling (MDS) \cite{ref27}, as well as supervised methods such as linear discriminant analysis (LDA) \cite{ref28}, supervised locality preserving projections (SLPP) \cite{ref26}, and neighborhood component analysis (NCA) \cite{ref29}. Among the various methods, Mahalanobis-based metrics \cite{ref10, ref30, ref31, ref32, ref33, ref34} are most relevant to our work. The likelihood ratio-based metric used in our proposed GLRTML under the assumption of multivariate Gaussian distributions can be interpreted as a Mahalanobis metric. However, existing Mahalanobis metric methods \cite{ref10, ref30, ref31, ref32} primarily focus on learning a static metric space from the data without using the Mahalanobis distance as feedback to optimize the feature extraction process. This separation limits their ability to adapt the metric space based on the learned feature representations, potentially leading to suboptimal performance during inference. In contrast, GLRTML constructs a loss function based on GLRT that optimizes neural network parameters and updates the metric space in an end-to-end manner. By utilizing global data distribution information during both training and inference stages, GLRTML enhances the discriminative power of the metric space, resulting in superior inference capabilities. 

In the era of deep learning, metric learning mainly revolves around the design of loss functions and training strategies \cite{ref19}. Loss functions are generally classified into two types: sample-to-class distance-based loss \cite{ref21, ref22, ref33, ref34, ref35} and sample-to-sample distance-based loss \cite{ref36, ref37, ref38, ref39, ref40}. Sample-to-class distance-based loss frames CBRSIR training as an image classification problem. For example, identity loss \cite{ref33} uses the cross-entropy function to predict class probabilities over a SoftMax layer, which is then removed in the testing phase to extract embeddings. UCE loss \cite{ref35} enforces a unified threshold to ensure that all positive sample-to-class distances are smaller than all negative sample-to-class distances. On the other hand, sample-to-sample distance-based loss \cite{ref36, ref37, ref38, ref39, ref40} constructs positive/negative sample pairs and trains network parameters to minimize the intra-class distance and maximize the inter-class distance. Contrastive loss \cite{ref36} and triplet loss \cite{ref37} construct positive/negative sample pairs in binary and triplet form, respectively. Sun et al. \cite{ref41} unify sample-to-class and sample-to-sample loss perspectives by reweighting pairs based on their similarity scores, emphasizing less optimized pairs. While these approaches introduce well-designed loss functions to improve the network's embedding capabilities, they fail to account for global data relationships, often resulting in overfitting to simple positive/negative sample pairs due to GPU memory limitations. 

In addition to designing loss functions, several metric learning approaches have attempted to address the lack of global information exploitation that leads to insufficient attention to hard samples by employing specific training strategies. For example, identity sampling \cite{ref42} aims to better represent the global data distribution within each batch by randomly selecting classes and sampling a fixed number of images from each class, thereby addressing class imbalance. However, this method only approximates the global distribution at the batch level and may not capture the full complexity of data relationships. Hard sample mining \cite{ref37} focuses on the hardest positive and negative pairs in the triplet loss, thereby directing the network's attention to more challenging samples. Similarly, sparse pairwise (SP) loss \cite{ref40} introduces an adaptive positive mining strategy that adapts to intra-class variation by dynamically selecting the most informative triplets within each batch. However, hard sample mining and SP loss rely on local sample relationships within batches and do not fully exploit global data structures, potentially missing important global relationships. Center loss \cite{ref21} and class-wise center banks \cite{ref22} attempt to incorporate global data relationships through class center embeddings. While these methods encourage networks to consider global distributions, they often reduce intra-class variability by averaging over class samples. This simplification can lead to a focus on easier sample pairs that stabilize during training, reducing the model's capacity for further optimization and potentially leading to overfitting. Although these strategies result in neural networks focusing on global data relationships or hard samples to some extent, they typically rely on complex rule-based designs and are limited by batch-level operations. As a result, they may not effectively capture the full global data distribution or adequately address hard samples.

In contrast, our proposed GLRTML method innovatively incorporates global statistical information from the entire training data into the training phase, allowing the model to estimate the difficulty of each sample pair. This guides the network to focus on more difficult samples, preventing overfitting and improving generalization. By incorporating the Neyman-Pearson theorem \cite{ref56}, we ensure the optimality of the learned metric space when the probability distributions are accurately estimated. Since the metric space and the neural network are trained alternately, the learned embeddings achieve asymptotic optimality within the metric space.

\subsection{Pseudo-Label-Based Unsupervised Domain Adaptation}
Pseudo-label-based UDA methods have demonstrated excellent performance by allowing models to adapt to new domains without extensive labeled data. These methods generally follow a two-step training process: (1) supervised pre-training on the source domain, and (2) unsupervised fine-tuning on the target domain. Pseudo-labels are generated either by clustering embeddings \cite{ref6, ref17, ref49, ref51, ref52} or by measuring similarities with exemplar embeddings \cite{ref53, ref54}. Clustering-based methods \cite{ref6, ref17, ref49, ref51, ref52} have achieved SOTA results. For example, Sun et al. \cite{ref6} propose an unsupervised deep hashing method based on soft pseudo-labels, which uses soft pseudo-labels to more accurately capture global similarities between images. Hou et al. \cite{ref17} introduce a pseudo-label consistency learning-based UDA method, where pseudo-label self-training and consistency regularization minimize the discrepancy between the target domain and its perturbed outputs.

Inspired by the effectiveness of these pseudo-label-based UDA methods, we adopt a similar approach to address the statistical distribution parameters estimation problem in our GLRTML under domain shift scenarios. However, our proposed CPLFPA method offers a significant advantage: it does not require iterative fine-tuning of a large number of neural network parameters. Instead, CPLFPA efficiently adjusts a few statistical model parameters, resulting in a faster and more computationally efficient adaptation process. This efficiency is particularly important for CBRSOR tasks, where rapid adaptation to new domains is critical.

\section{Method}

In this section, we provide a detailed description and analysis about the proposed GLRTML and CPLFPA, respectively.
\subsection{Generalized Likelihood Ratio Test-Based Metric Learning}
Our proposed GLRTML models the CBRSOR problem as a hypothesis testing problem. The case of the unpaired query image and gallery image is defined as null hypothesis $H_0$. Correspondingly, the case of the paired query image and gallery image is defined as alternative hypothesis $H_1$. For any query image and gallery image pair $(i,j)$, the likelihood ratio between the alternative hypothesis distribution $p(\cdot | H_1)$ and null hypothesis distribution $p(\cdot | H_0)$ is used to define the similarity $s(i, j)$ between query image $I_i$ and gallery image $I_j$:
\begin{equation}
    s(i, j) = \log \left( \frac{p(I_i, I_j | H_1)}{p(I_i, I_j | H_0)} \right). \tag{1}
\end{equation}
Thanks to the powerful representation ability of neural networks, we can calculate the ratio in embeddings space instead of the original images space:
\begin{equation}
    s(i, j) = \log \left( \frac{p(\boldsymbol{x}_{\boldsymbol{\theta}, i}, \boldsymbol{x}_{\boldsymbol{\theta}, j} | H_1)}{p(\boldsymbol{x}_{\boldsymbol{\theta}, i}, \boldsymbol{x}_{\boldsymbol{\theta}, j} | H_0)} \right), \tag{2}
\end{equation}
where $\boldsymbol{x}_{\boldsymbol{\theta}, i}$ and $\boldsymbol{x}_{\boldsymbol{\theta}, j}$ represent the embeddings of images $I_i$ and $I_j$ obtained through inference of the neural network with parameter $\boldsymbol{\theta}$. From a statistical inference perspective, a high value of $s(i, j)$ means accepting hypothesis $H_1$ with high confidence, i.e., believing that $I_i$ and $I_j$ are similar. On the contrary, a low value of $s(i, j)$ means $I_i$ and $I_j$ are not similar. Considering that $s(i, j)$ should be equivalent to $s(j, i)$, we replace $\boldsymbol{x}_{\boldsymbol{\theta}, i}$ and $\boldsymbol{x}_{\boldsymbol{\theta}, j}$ with differential embedding $\boldsymbol{x}_{\boldsymbol{\theta}, ij} = \boldsymbol{x}_{\boldsymbol{\theta}, i} - \boldsymbol{x}_{\boldsymbol{\theta}, j}$ rather than cascading embedding, thereby eliminating the bias caused by the position of elements in embedding and introducing symmetry:

\begin{equation}
    s(i, j) = \log \left( \frac{p(\boldsymbol{x}_{\boldsymbol{\theta}, ij} | H_1)}{p(\boldsymbol{x}_{\boldsymbol{\theta}, ij} | H_0)} \right). \tag{3}
\end{equation}

However, the alternative hypothesis distribution $p(\boldsymbol{x}_{\boldsymbol{\theta}, ij} | H_1)$ and null hypothesis distribution $p(\boldsymbol{x}_{\boldsymbol{\theta}, ij} | H_0)$ are often difficult to obtain in advance, so they need to be calculated through parametric estimation. Let $\hat{p}(\boldsymbol{x}_{\boldsymbol{\theta}, ij} | \boldsymbol{\vartheta}_1)$ and $\hat{p}(\boldsymbol{x}_{\boldsymbol{\theta}, ij} | \boldsymbol{\vartheta}_0)$ denote the parameterized estimates of $p(\boldsymbol{x}_{\boldsymbol{\theta}, ij} | H_1)$ and $p(\boldsymbol{x}_{\boldsymbol{\theta}, ij} | H_0)$ respectively, i.e., estimated distribution, where $\boldsymbol{\vartheta}_1$ and $\boldsymbol{\vartheta}_0$ are estimated distribution parameters. Multivariate Gaussian and Gaussian mixture models (GMM) \cite{ref58} are used to model distributions respectively, and we obtain multivariate Gaussian-GLRTML (MG-GLRTML) and Gaussian mixture models-GLRTML (GMM-GLRTML), which are introduced as follows.

\begin{figure}[!t]
\centering
\includegraphics[width=4.0in]{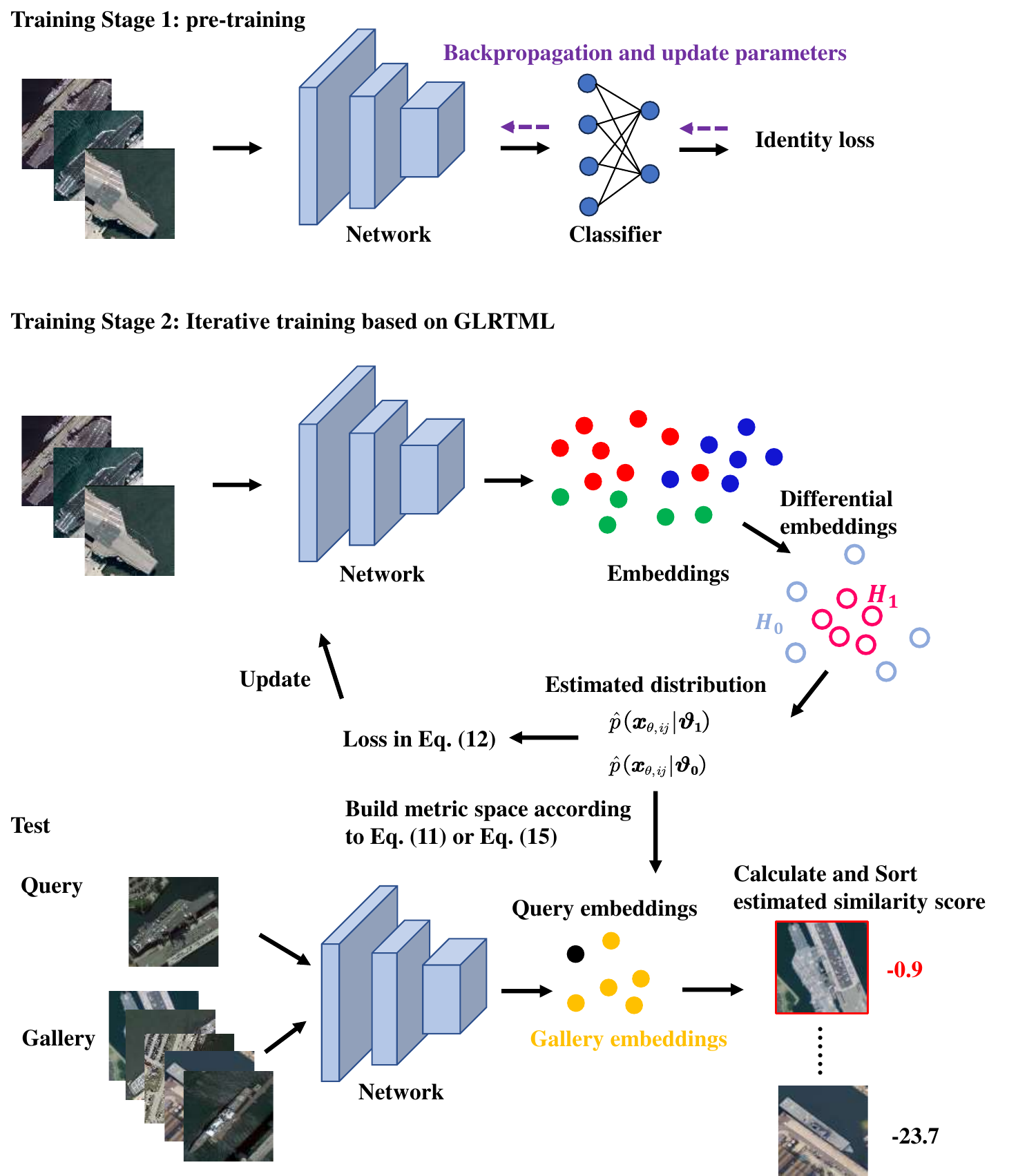}
\caption{The flow chart of proposed GLRTML.}
\label{fig_2}
\end{figure}

\subsubsection*{\bf 1) MG-GLRTML}

We assume that $\hat{p}(\boldsymbol{x}_{\boldsymbol{\theta}, ij} | \boldsymbol{\vartheta}_1)$ and $\hat{p}(\boldsymbol{x}_{\boldsymbol{\theta}, ij} | \boldsymbol{\vartheta}_0)$ are both composed of a $d$-dimensional multivariate Gaussian distribution with mean vector $\boldsymbol{\mu}_1$, $\boldsymbol{\mu}_0$ and covariance matrix $\boldsymbol{\Sigma}_1$, $\boldsymbol{\Sigma}_0$:

\begin{align}
    \hat{p}(\boldsymbol{x}_{\boldsymbol{\theta}, ij} | \boldsymbol{\vartheta}_1) &= \hat{p}(\boldsymbol{x}_{\boldsymbol{\theta}, ij} | \boldsymbol{\mu}_1, \boldsymbol{\Sigma}_1) \notag \\
    &= \frac{1}{(2\pi)^{d/2} |\boldsymbol{\Sigma}_1|^{1/2}} \notag \\
    &\quad \times \exp \left( 
    -\frac{1}{2} (\boldsymbol{x}_{\boldsymbol{\theta}, ij} - \boldsymbol{\mu}_1)^\top \boldsymbol{\Sigma}_1^{-1} (\boldsymbol{x}_{\boldsymbol{\theta}, ij} - \boldsymbol{\mu}_1) 
    \right). \tag{4}
\end{align}

\begin{align}
    \hat{p}(\boldsymbol{x}_{\boldsymbol{\theta}, ij} | \boldsymbol{\vartheta}_0) &= \hat{p}(\boldsymbol{x}_{\boldsymbol{\theta}, ij} | \boldsymbol{\mu}_0, \boldsymbol{\Sigma}_0) \notag \\
    &= \frac{1}{(2\pi)^{d/2} |\boldsymbol{\Sigma}_0|^{1/2}} \notag \\
    &\quad \times \exp \left( 
    -\frac{1}{2} (\boldsymbol{x}_{\boldsymbol{\theta}, ij} - \boldsymbol{\mu}_0)^\top \boldsymbol{\Sigma}_0^{-1} (\boldsymbol{x}_{\boldsymbol{\theta}, ij} - \boldsymbol{\mu}_0) 
    \right). \tag{5}
\end{align}
We substitute Eq.~(4) and Eq.~(5) into Eq.~(3) respectively, replacing $p(\cdot | H_1)$ and $p(\cdot | H_0)$:
\begin{align}
    s(i, j) &= \frac{1}{2} (\boldsymbol{x}_{\boldsymbol{\theta}, ij} - \boldsymbol{\mu}_0)^\top \boldsymbol{\Sigma}_0^{-1} (\boldsymbol{x}_{\boldsymbol{\theta}, ij} - \boldsymbol{\mu}_0) \notag \\
            &\quad - \frac{1}{2} (\boldsymbol{x}_{\boldsymbol{\theta}, ij} - \boldsymbol{\mu}_1)^\top \boldsymbol{\Sigma}_1^{-1} (\boldsymbol{x}_{\boldsymbol{\theta}, ij} - \boldsymbol{\mu}_1) + C_0. \tag{6}
\end{align}
Where $C_0$ is a statistically irrelevant part. Considering that $s(i, j)$ is ultimately used for sorting, and note that the symmetry of probability space (i.e., $s(i, j)$ should be equivalent to $s(j, i)$), we assume that differential embedding is zero-mean, we further simplify the Eq.~(6) and obtain the similarity score for MG-GLRTML:

\begin{equation}
    s(i, j) = \boldsymbol{x}_{\boldsymbol{\theta}, ij}^\top \left( \boldsymbol{\Sigma}_0^{-1} - \boldsymbol{\Sigma}_1^{-1} \right) \boldsymbol{x}_{\boldsymbol{\theta}, ij}. \tag{7}
\end{equation}
Eq.~(7) reveals that our metric space is in the form of Mahalanobis distance. Further, we use maximum likelihood estimation (MLE) \cite{ref57} to estimate the covariance matrix $\boldsymbol{\Sigma}_1$ and $\boldsymbol{\Sigma}_0$, and we assume that there is a total of $N_1$ and $N_0$ available image pairs in the training set for hypothesis $H_1$ set $X_1$ and hypothesis $H_0$ set $X_0$, respectively. For ease of expression, we denote the differential embeddings of $l$-th positive pair or negative pair as $\boldsymbol{x}_{\boldsymbol{\theta}, l}$:

\noindent
\begin{equation}
    L(\boldsymbol{\Sigma}_1) = \prod_{l=0}^{N_1 - 1} p(\boldsymbol{x}_{\boldsymbol{\theta}, l}^{+} | \boldsymbol{\Sigma}_1), \quad \boldsymbol{x}_{\boldsymbol{\theta}, l}^{+} \in X_1. \tag{8}
\end{equation}
Let $\frac{\partial}{\partial \boldsymbol{\Sigma}_1} \log L(\boldsymbol{\Sigma}_1) = 0$, and we obtain the estimate of $\boldsymbol{\Sigma}_1$:
\begin{equation}
    \hat{\boldsymbol{\Sigma}}_1 = \frac{1}{N_1} \sum_{l=0}^{N_1 - 1} (\boldsymbol{x}_{\boldsymbol{\theta}, l}^{+})(\boldsymbol{x}_{\boldsymbol{\theta}, l}^{+})^\top, \quad \boldsymbol{x}_{\boldsymbol{\theta}, l}^{+} \in X_1. \tag{9}
\end{equation}
In the same way, the estimate of $\boldsymbol{\Sigma}_0$ is as follows:
\begin{equation}
    \hat{\boldsymbol{\Sigma}}_0 = \frac{1}{N_0} \sum_{l=0}^{N_0 - 1} (\boldsymbol{x}_{\boldsymbol{\theta}, l}^{-})(\boldsymbol{x}_{\boldsymbol{\theta}, l}^{-})^\top, \quad \boldsymbol{x}_{\boldsymbol{\theta}, l}^{-} \in X_0. \tag{10}
\end{equation}
Therefore, rewrite Eq.~(7) to obtain the estimated similarity score:
\begin{equation}
    \hat{s}(i, j) = \boldsymbol{x}_{\boldsymbol{\theta}, ij}^\top \left( \hat{\boldsymbol{\Sigma}}_0^{-1} - \hat{\boldsymbol{\Sigma}}_1^{-1} \right) \boldsymbol{x}_{\boldsymbol{\theta}, ij}. \tag{11}
\end{equation}
We clip the spectrum of $\hat{\boldsymbol{\Sigma}}_0^{-1} - \hat{\boldsymbol{\Sigma}}_1^{-1}$ by eigen analysis to make sure $\hat{\boldsymbol{\Sigma}}_0^{-1} - \hat{\boldsymbol{\Sigma}}_1^{-1}$ is positive definite. In order to train neural network end-to-end, we construct a loss function based on the estimated similarity score. The baseline of our loss function is the weighted sum of unified loss~\cite{ref41} and identity loss~\cite{ref33}. We replace the cosine similarity in unified loss with estimated similarity score and get the following loss function:

\begin{align}
    \mathcal{L} &= \mathcal{L}_{\text{GLRTML}} + \alpha \mathcal{L}_{\text{Identity}} \notag \\
    &= \log \left[ 1 + \sum_{m=0}^{M_1 - 1} \sum_{l=0}^{M_0 - 1} \exp \left( \nu \left( \hat{s}_n^{l} - \hat{s}_p^{m} \right) \right) \right] \notag \\
    &\quad - \alpha \frac{1}{M} \sum_{i=0}^{M - 1} \log \left( \tilde{p}(y_i | \boldsymbol{x}_{\boldsymbol{\theta}, i}) \right) \tag{12} \\
    &= \log \left[ 1 + \sum_{l=0}^{M_0 - 1} \exp \left( \nu \hat{s}_n^{l} \right) \sum_{m=0}^{M_1 - 1} \exp \left( -\nu \hat{s}_p^{m} \right) \right] \notag \\
    &\quad - \alpha \frac{1}{M} \sum_{i=0}^{M - 1} \log \left( \tilde{p}(y_i | \boldsymbol{x}_{\boldsymbol{\theta}, i}) \right), \notag
\end{align}
where $M_1$, $M_0$, and $M$ respectively represent the total number of positive sample pairs, negative sample pairs, and images within a batch. $\hat{s}_p^{m}$ and $\hat{s}_n^{l}$ are the estimated similarity scores of positive sample pairs and negative sample pairs, respectively. The indices $m$, $l$, and $i$ index positive sample pairs, negative sample pairs, and images within the batch, respectively. The variable $y_i$ is the class label of the image embedding $\boldsymbol{x}_{\boldsymbol{\theta}, i}$, and $\tilde{p}(y_i | \boldsymbol{x}_{\boldsymbol{\theta}, i})$ represents the predicted probability of the embedding $\boldsymbol{x}_{\boldsymbol{\theta}, i}$ being classified as class $y_i$ by the classifier. The parameter $\nu$ is a temperature coefficient used to control the scale of the exponential terms, and $\alpha$ is a weighting coefficient that balances the GLRTML loss $\mathcal{L}_{\text{GLRTML}}$ and the identity loss $\mathcal{L}_{\text{Identity}}$.

Calculating the loss function in Eq.~(12) requires first completing the estimation of the covariance matrix in Eq.~(9) and Eq.~(10). After the calculation of the loss function is completed, due to the update of the neural network parameters, the distribution of embeddings changes, and the statistical distribution parameters (i.e., the covariance matrix) need to be re-estimated. Therefore, we design an iterative training strategy to collaboratively update network parameters and statistical distribution parameters during training. In addition, we design a two-stage training strategy, i.e., first using identity loss~\cite{ref33}'s label learning ability to obtain a pre-trained model, and then training based on Eq.~(12).

In the test phase, the embeddings of query image and gallery image are obtained through the trained neural network, and then calculate the estimated similarity score according to Eq.~(11), where $\hat{\boldsymbol{\Sigma}}_1$ and $\hat{\boldsymbol{\Sigma}}_0$ are estimated on the training set after the last epoch. Finally, we sort the estimated similarity score in descending order to get the retrieval results. Algorithm~1 and Fig.~(2) show the specific process.

\begin{algorithm}
\caption{MG-GLRTML}
\label{alg:mg-glrtml}

\textbf{Training:}

\textbf{Input:} \\
Images set $\{I_i\}$ and corresponding label set $\{y_i\}$ from training set. Neural network $\phi$ with learnable parameters $\boldsymbol{\theta}$. \\

\textbf{Training Stage 1:}
\begin{algorithmic}[1]
    \For{each epoch $t = 0, \dots, T_0 - 1$}
        \For{each batch $b = 0, \dots, B - 1$ and sample $\{I_i, y_i\}_b$}
            \State Forward propagations $\boldsymbol{x}_{\boldsymbol{\theta}_{t,b}, i} = \phi(I_i, \boldsymbol{\theta}_{t,b})$, and batch gradient descent \cite{ref59} based on identity loss \cite{ref33} to update parameters.
        \EndFor
    \EndFor
\end{algorithmic}

\textbf{Training Stage 2:}
\begin{algorithmic}[1]
    \For{each epoch $t = T_0, \dots, T_1 - 1$}

        \State Forward propagations $\boldsymbol{x}_{\boldsymbol{\theta}_{t, 0}, i} = \phi(I_i, \boldsymbol{\theta}_{t, 0})$.

        \State Calculate statistical distribution parameters $\bm{\hat{\Sigma}}_1$ and $\bm{\hat{\Sigma}}_0$ based on Eq.~(9) and Eq.~(10).
        \For{each batch $b = 0, \dots, B - 1$ and sample $\{I_i, y_i\}_b$}
            \State Forward propagations $\boldsymbol{x}_{\boldsymbol{\theta}_{t, b}, i} = \phi(I_i, \boldsymbol{\theta}_{t, b})$, and batch gradient descent \cite{ref59} based on the loss defined in Eq.~(12) to obtain updated parameters.
        \EndFor
    \EndFor
    \State Calculate statistical distribution parameters $\bm{\hat{\Sigma}}_1$ and $\bm{\hat{\Sigma}}_0$ based on Eq.~(9) and Eq.~(10).
\end{algorithmic}

\textbf{Test:}

\textbf{Input:} \\
Images set $\{I_i\}$ and $\{I_j\}$ from query set and gallery set. Trained neural network $\phi$ with parameters $\boldsymbol{\theta}_{T-1, B-1}$, and statistical distribution parameters $\bm{\hat{\Sigma}}_1$ and $\bm{\hat{\Sigma}}_0$ obtained from training. \\

\begin{algorithmic}[1]
    \For{each image $I_i$ in query set}
        \State Forward propagation: $\boldsymbol{x}_{i} = \phi(I_i, \boldsymbol{\theta}_{T-1, B-1})$
        \For{each image $I_j$ in gallery set}
            \State Forward propagation: $\boldsymbol{x}_{j} = \phi(I_j, \boldsymbol{\theta}_{T-1, B-1})$
            \State Calculate differential embedding $\boldsymbol{x}_{\boldsymbol{\theta}_{T-1, B-1},ij}$.
            \State Calculate estimated similarity score using Eq.~(11)
        \EndFor
        \State Sort estimated similarity scores in descending order to get the retrieval results for $I_i$
    \EndFor

\end{algorithmic}
\end{algorithm}

\subsubsection*{\bf 2) GMM-GLRTML}

We assume that $\hat{p}(\boldsymbol{x}_{\boldsymbol{\theta}, ij}|\boldsymbol{\vartheta}_1)$ and $\hat{p}(\boldsymbol{x}_{\boldsymbol{\theta}, ij}|\boldsymbol{\vartheta}_0)$ are composed of $K_1$ and $K_0$ $d$-dimensional multivariate Gaussian distribution, respectively:

\begin{align}
    \hat{p}(\boldsymbol{x}_{\boldsymbol{\theta}, ij}|\boldsymbol{\vartheta}_1) &= \sum_{k=0}^{K_1-1} \pi_k^1 \mathcal{N}(\boldsymbol{x}_{\boldsymbol{\theta}, ij} | \boldsymbol{\mu}_k^1, \boldsymbol{\Sigma}_k^1), \quad \text{s.t.} \quad \sum_{k=0}^{K_1-1} \pi_k^1 = 1, \tag{13}
\end{align}

\begin{align}
    \hat{p}(\boldsymbol{x}_{\boldsymbol{\theta}, ij}|\boldsymbol{\vartheta}_0) &= \sum_{k=0}^{K_0-1} \pi_k^0 \mathcal{N}(\boldsymbol{x}_{\boldsymbol{\theta}, ij} | \boldsymbol{\mu}_k^0, \boldsymbol{\Sigma}_k^0), \quad \text{s.t.} \quad \sum_{k=0}^{K_0-1} \pi_k^0 = 1. \tag{14}
\end{align}
Where $\pi_k^1$ and $\pi_k^0$ are weights of the $k$-th Gaussian component of hypothesis $H_1$ distribution and hypothesis $H_0$ distribution, respectively. $\mathcal{N}(\cdot)$ means Gaussian distribution, $\boldsymbol{\mu}_k^1$ and $\boldsymbol{\Sigma}_k^1$ are mean vector and covariance matrix of the $k$-th Gaussian component of hypothesis $H_1$ distribution. $\boldsymbol{\mu}_k^0$ and $\boldsymbol{\Sigma}_k^0$ are mean vector and covariance matrix of the $k$-th Gaussian component of hypothesis $H_0$ distribution. Here, $\boldsymbol{\vartheta}_1 = (\pi_k^1, \boldsymbol{\mu}_k^1, \boldsymbol{\Sigma}_k^1)_{k=1, \dots, K_1}$ and $\boldsymbol{\vartheta}_0 = (\pi_k^0, \boldsymbol{\mu}_k^0, \boldsymbol{\Sigma}_k^0)_{k=1, \dots, K_0}$, the estimated similarity score of GMM-GLRTML is:

\begin{equation}
    \hat{s}(i, j) = \log \left( \frac{\hat{p}(\boldsymbol{x}_{\boldsymbol{\theta}, ij}|\boldsymbol{\vartheta}_1)}{\hat{p}(\boldsymbol{x}_{\boldsymbol{\theta}, ij}|\boldsymbol{\vartheta}_0)} \right). \tag{15}
\end{equation}

We use expectation-maximization (EM) algorithm \cite{ref58} to calculate the estimated distribution parameters $\boldsymbol{\vartheta}_1$ and $\boldsymbol{\vartheta}_0$. We denote the differential embeddings of $l$-th positive pair or negative pair as $\boldsymbol{x}_{\boldsymbol{\theta}, l}$, and denote hypothesis $H_1$ set as $X_1$ with $N_1$ available image pairs, hypothesis $H_0$ set as $X_0$ with $N_0$ available image pairs. Let $\mathbf{Z}^1 = \{z_0^1, \dots, z_{N_1 - 1}^1 \}$ and $\mathbf{Z}^0 = \{z_0^0, \dots, z_{N_0 - 1}^0\}$ denote latent variable, i.e., indicator variable $z_l^1 = \mathbf{e}_k$ means that the sample $\boldsymbol{x}_{\boldsymbol{\theta}, l}$ comes from the $k$-th Gaussian component in hypothesis $H_1$ distribution, where $\mathbf{e}_k$ represents a vector whose $k$-th element is 1 and other elements are 0. For the $s$-th iteration of the EM algorithm, the expectation of log-likelihood function is:

\begin{align}
    Q(\boldsymbol{\vartheta}_1|\boldsymbol{\vartheta}_1^{(s)}) &= 
    \mathbb{E}_{\mathbf{Z}^1 | X_1, \boldsymbol{\vartheta}_1^{(s)}} 
    \left[ \log \hat{p}(\boldsymbol{x}_{\boldsymbol{\theta}, ij}, \mathbf{Z}^1 | \boldsymbol{\vartheta}_1) \right] \notag \\
    &= \mathbb{E}_{\mathbf{Z}^1 | X_1, \boldsymbol{\vartheta}_1^{(s)}} 
    \Bigg[ \sum_{l=0}^{N_1 - 1} \sum_{k=0}^{K_1 - 1} z_{l,k}^1 &\Big( \log \pi_k^1 \notag \\
    &\quad + \log \mathcal{N}(\boldsymbol{x}_{\boldsymbol{\theta}, l} | \boldsymbol{\mu}_k^1, \boldsymbol{\Sigma}_k^1) \Big) 
    \Bigg], \tag{16}
\end{align}

\begin{align}
    Q(\boldsymbol{\vartheta}_0|\boldsymbol{\vartheta}_0^{(s)}) &= 
    \mathbb{E}_{\mathbf{Z}^0 | X_0, \boldsymbol{\vartheta}_0^{(s)}} 
    \left[ \log \hat{p}(\boldsymbol{x}_{\boldsymbol{\theta}, ij}, \mathbf{Z}^0 | \boldsymbol{\vartheta}_0) \right] \notag \\
    &= \mathbb{E}_{\mathbf{Z}^0 | X_0, \boldsymbol{\vartheta}_0^{(s)}} 
    \Bigg[ \sum_{l=0}^{N_0 - 1} \sum_{k=0}^{K_0 - 1} z_{l,k}^0 &\Big( \log \pi_k^0 \notag \\
    &\quad + \log \mathcal{N}(\boldsymbol{x}_{\boldsymbol{\theta}, l} | \boldsymbol{\mu}_k^0, \boldsymbol{\Sigma}_k^0) \Big) 
    \Bigg]. \tag{17}
\end{align}

Calculate the posterior probability that each sample comes from each component given $\boldsymbol{\vartheta}_1$ and $\boldsymbol{\vartheta}_0$, which is called expectation step (E-step). The E-step is as follows:

\begin{align}
    \gamma_{l,k}^1 &= \mathbb{E}[z_{l,k} | X_1, \boldsymbol{\vartheta}_1] \notag \\
    &= \frac{\pi_k^1 \mathcal{N}(\boldsymbol{x}_{\boldsymbol{\theta}, l} | \boldsymbol{\mu}_k^1, \boldsymbol{\Sigma}_k^1)}{\sum_{i=0}^{K_1 - 1} \pi_i^1 \mathcal{N}(\boldsymbol{x}_{\boldsymbol{\theta}, l} | \boldsymbol{\mu}_i^1, \boldsymbol{\Sigma}_i^1)}, \quad \boldsymbol{x}_{\boldsymbol{\theta}, l}^+ \in X_1, \tag{18}
\end{align}

\begin{align}
    \gamma_{l,k}^0 &= \mathbb{E}[z_{l,k} | X_0, \boldsymbol{\vartheta}_0] \notag \\
    &= \frac{\pi_k^0 \mathcal{N}(\boldsymbol{x}_{\boldsymbol{\theta}, l} | \boldsymbol{\mu}_k^0, \boldsymbol{\Sigma}_k^0)}{\sum_{i=0}^{K_0 - 1} \pi_i^0 \mathcal{N}(\boldsymbol{x}_{\boldsymbol{\theta}, l} | \boldsymbol{\mu}_i^0, \boldsymbol{\Sigma}_i^0)}, \quad \boldsymbol{x}_{\boldsymbol{\theta}, l}^- \in X_0. \tag{19}
\end{align}

In maximization step (M-step), we update the estimated distribution parameters $\boldsymbol{\vartheta}_1$ and $\boldsymbol{\vartheta}_0$ so that the likelihood function is maximized given the posterior probability:

\begin{align}
    \max Q(\boldsymbol{\vartheta}_1|\boldsymbol{\vartheta}_1^{(s)}) &= \sum_{l=0}^{N_1 - 1} \sum_{k=0}^{K_1 - 1} \gamma_{l,k}^1 \Big( \log \pi_k^1 \notag \\
    &\quad + \log \mathcal{N}(\boldsymbol{x}_{\boldsymbol{\theta}, l}^+ | \boldsymbol{\mu}_k^1, \boldsymbol{\Sigma}_k^1) \Big) \tag{20} \\[10pt]
    \text{s.t.} &\quad \sum_{k=0}^{K_1-1} \pi_k^1 = 1. \notag
\end{align}

The analytical solution to the above problem is:

\begin{equation}
\begin{cases}
    \pi_k^{1, (s+1)} = \frac{1}{N_1} \sum_{l=0}^{N_1 - 1} \gamma_{l,k}^1 \\[10pt]
    \boldsymbol{\mu}_k^{1, (s+1)} = \frac{\sum_{l=0}^{N_1 - 1} \gamma_{l,k}^1 \boldsymbol{x}_{\boldsymbol{\theta}, l}^+}{\sum_{l=0}^{N_1 - 1} \gamma_{l,k}^1} \\[10pt]
    \boldsymbol{\Sigma}_k^{1, (s+1)} = \frac{\sum_{l=0}^{N_1 - 1} \gamma_{l,k}^1 
    \left( \boldsymbol{x}_{\boldsymbol{\theta}, l}^+ - \boldsymbol{\mu}_k^{1, (s+1)} \right) 
    \left( \boldsymbol{x}_{\boldsymbol{\theta}, l}^+ - \boldsymbol{\mu}_k^{1, (s+1)} \right)^\top}
    {\sum_{l=0}^{N_1 - 1} \gamma_{l,k}^1}
\end{cases} \tag{21}
\end{equation}

In the same way, we can get the updated $\boldsymbol{\vartheta}_0$:

\begin{equation}
\begin{cases}
    \pi_k^{0, (s+1)} = \frac{1}{N_0} \sum_{l=0}^{N_0 - 1} \gamma_{l,k}^0 \\[10pt]
    \boldsymbol{\mu}_k^{0, (s+1)} = \frac{\sum_{l=0}^{N_0 - 1} \gamma_{l,k}^0 \boldsymbol{x}_{\boldsymbol{\theta}, l}^-}{\sum_{l=0}^{N_0 - 1} \gamma_{l,k}^0} \\[10pt]
    \boldsymbol{\Sigma}_k^{0, (s+1)} = \frac{\sum_{l=0}^{N_0 - 1} \gamma_{l,k}^0 
    \left( \boldsymbol{x}_{\boldsymbol{\theta}, l}^- - \boldsymbol{\mu}_k^{0, (s+1)} \right) 
    \left( \boldsymbol{x}_{\boldsymbol{\theta}, l}^- - \boldsymbol{\mu}_k^{0, (s+1)} \right)^\top}
    {\sum_{l=0}^{N_0 - 1} \gamma_{l,k}^0}
\end{cases} \tag{22}
\end{equation}

By repeatedly executing E-step and M-step until the estimated distribution parameters $\boldsymbol{\vartheta}_1$ and $\boldsymbol{\vartheta}_0$ converge, we can obtain the maximum likelihood estimate of $\boldsymbol{\vartheta}_1$ and $\boldsymbol{\vartheta}_0$. In the same way as MG-GLRTML, we get the same form of loss function as Eq.~(12) to train the neural network end-to-end. The only difference for loss function is that the estimated similarity score here is based on Eq.~(15). Similar to MG-GLRTML, we design an iterative training process to collaboratively train the neural network and update statistical distribution parameters. Besides, we follow the two-stage training strategy and testing process in MG-GLRTML. Algorithm 2 and Fig.~(2) show the specific process of GMM-GLRTML.

\begin{algorithm}
\caption{GMM-GLRTML}
\label{alg:gmm-glrtml}

\textbf{Training:}

\textbf{Input:} \\
Images set $\{I_i\}$ and corresponding label set $\{y_i\}$ from training set. Neural network $\phi$ with learnable parameters $\boldsymbol{\theta}$. \\

\textbf{Training Stage 1:}
\begin{algorithmic}[1]
    \For{each epoch $t = 0, \ldots, T_0 - 1$}
        \For{each batch $b = 0, \ldots, B - 1$ and sample $\{I_i, y_i\}_b$}
            \State Forward propagations $\boldsymbol{x}_{\boldsymbol{\theta}, b, i} = \phi(I_i, \boldsymbol{\theta}_{t, b})$, and batch gradient descent \cite{ref59} based on identity loss \cite{ref33} to update parameters.
        \EndFor
    \EndFor
\end{algorithmic}

\textbf{Training Stage 2:}
\begin{algorithmic}[1]
    \For{each epoch $t = T_0, \ldots, T_1 - 1$}
        \State Forward propagations $\boldsymbol{x}_{\boldsymbol{\theta}, 0, i} = \phi(I_i, \boldsymbol{\theta}_{t, 0})$.
        \State Calculate statistical distribution parameters $\boldsymbol{\vartheta}_1$ and $\boldsymbol{\vartheta}_0$ based on EM algorithm \cite{ref58}, i.e., Eq.~(18), Eq.~(19), Eq.~(21), and Eq.~(22).
        \For{each batch $b = 0, \ldots, B - 1$ and sample $\{I_i, y_i\}_b$}
            \State Forward propagations $\boldsymbol{x}_{\boldsymbol{\theta}, b, i} = \phi(I_i, \boldsymbol{\theta}_{t, b})$, and batch gradient descent \cite{ref59} based on the loss defined in Eq.~(12) to obtain updated parameters.
        \EndFor
    \EndFor
    \State Calculate statistical distribution parameters $\boldsymbol{\vartheta}_1$ and $\boldsymbol{\vartheta}_0$ based on EM algorithm \cite{ref58}, i.e., Eq.~(18), Eq.~(19), Eq.~(21), and Eq.~(22).
\end{algorithmic}

\textbf{Test:}

\textbf{Input:} \\
Images set $\{I_i\}_q$ and $\{I_j\}_g$ from query set and gallery set. Trained neural network $\phi$ with parameters $\boldsymbol{\theta}_{T_1 - 1, B - 1}$, and statistical distribution parameters $\boldsymbol{\vartheta}_1$ and $\boldsymbol{\vartheta}_0$ obtained from training. \\

\begin{algorithmic}[1]
    \For{each image $I_i$ in query set}
        \State Forward propagations $\boldsymbol{x}_{\boldsymbol{\theta}_{T_1 - 1, B - 1}, i} = \phi(I_i, \boldsymbol{\theta}_{T_1 - 1, B - 1})$.
        \For{each image $I_j$ in gallery set}
            \State Forward propagations $\boldsymbol{x}_{\boldsymbol{\theta}_{T_1 - 1, B - 1}, j} = \phi(I_j, \boldsymbol{\theta}_{T_1 - 1, B - 1})$.
            \State Calculate differential embedding $\boldsymbol{x}_{\boldsymbol{\theta}_{T_1 - 1, B - 1}, ij}$.
            \State Calculate estimated similarity score based on Eq.~(15).
        \EndFor
        \State Sort estimated similarity score in descending order to get the retrieval results for $I_i$.
    \EndFor
\end{algorithmic}

\end{algorithm}

\subsubsection*{\bf 3) Discussion about GLRTML}

First, we discuss the asymptotic optimality of GLRTML:

\textbf{Discussion 1}. If the estimates of the hypothesis $H_1$ and the hypothesis $H_0$ are accurate, the training set and the test set are identically distributed, then GLRTML is asymptotically optimal for the given false alarm rate. 

The corresponding sketch of the proof is given in Appendix A. Note that there are two strong assumptions in Discussion 1: (1) the training and test sets are identically distributed; (2) the embedding distribution obeys a multivariate Gaussian or GMM. Considering that assumption (1) may not hold in real situations, especially in the new class situation. Therefore, we design CPLFPA to quickly update the statistical distribution parameters so that the GLRTML adapts to the test set distribution. In addition, we conduct a detailed investigation and discussion of the rationality of assumption (2) in Section IV. 

Second, we have stated in the introduction that GLRTML prevents overfitting to trivial simple samples by introducing global data distribution, thereby improving the learning ability of global relationship, but this is only a qualitative analysis. Here we give a quantitative analysis from the perspective of the loss function:

\textbf{Discussion 2}. GLRTML focuses more on difficult sample pairs in the data rather than trivial simple sample pairs during the training stage. 

The related proof is outlined in Appendix A. As can be seen from the above discussion, GLRTML estimates the difficulty of sample pairs in the global data by introducing global statistical information about the data, which in turn makes the gradient of the loss function larger with respect to the difficult sample pairs. GLRTML successfully makes the difficult sample pairs more important during the training process and is not constrained by the GPU's memory.

Third, we give the relationship between GLRTML and Mahalanobis metric in Theorem 1: 

\textbf{Theorem 1}. The estimated similarity score in MG-GLRTML can be equivalent to the form of Mahalanobis distance, and the estimated similarity score in GMM-GLRTML can be approximated to the form of weighted Mahalanobis distances.

Appendix A contains the sketch of the proof. Considering that the Mahalanobis distance uses concise parameters to characterize the scale and correlation of the data, which in turn measures the distance between data points, Theorem 1 inspired the idea that we can achieve fast domain adaptation by updating the parameters of GLRTML.

\subsection{Clustering Pseudo-Labels-based Fast Parameter Adaptation}

\begin{figure*}[!t]
\centering
\includegraphics[width=7.2in]{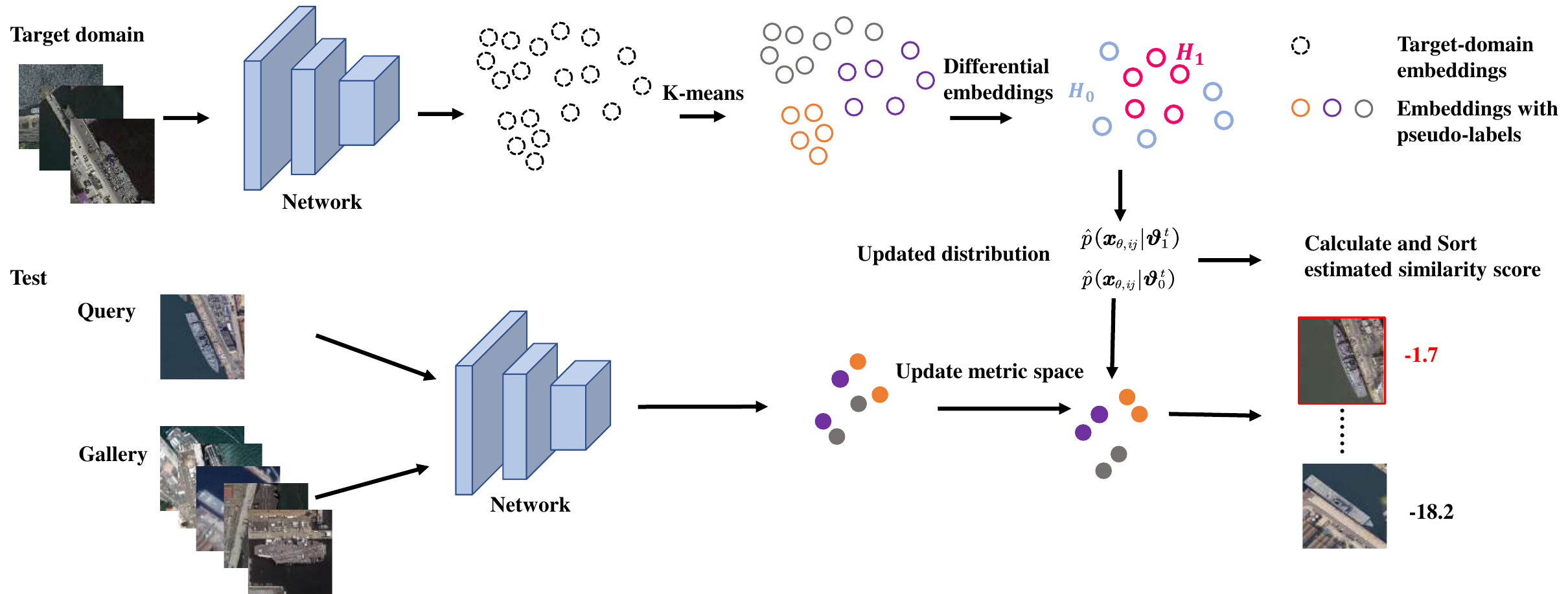}
\caption{The flow chart of proposed CPLFPA.}
\label{fig_3}
\end{figure*}

In order to efficiently solve the problem of GLRTML performance degradation caused by the domain difference between the training set and the test set, we propose CPLFPA. CPLFPA is based on the following two observations and analyses: (1) domain difference will cause the neural network to fail to extract well distributed embeddings (i.e., embeddings with compact intra-class and discrete inter-class embeddings) in the original metric space from test data. However, well distributed embeddings can be achieved by updating the metric space in addition to updating the neural network parameters. (2) The analysis in Discussion 1 inspires us that the correct estimation of distribution in test set is crucial to GLRTML, and the analysis in Theorem 1 reveals that we can update the metric space by updating $\boldsymbol{\vartheta}_1$  and $\boldsymbol{\vartheta}_0$, and the update is very fast. 

The flow chart of CPLFPA is shown in Fig.~(3). CPLFPA firstly extracts the embeddings of target-domain images set $\{I_i^t\}$ by the trained model:
\begin{equation}
    \boldsymbol{x}_{\boldsymbol{\theta}, i}^t = \phi(I_i^t, \tilde{\boldsymbol{\theta}}), \tag{23}
\end{equation}
where $\tilde{\boldsymbol{\theta}} = \boldsymbol{\theta}_{T_1-1, B-1}$, which means the pre-trained neural network parameters from GLRTML. We perform the K-means algorithm \cite{ref60} on the target-domain embeddings to obtain target-domain pseudo-labels $\{y_i^p\}$:
\begin{equation}
    \{y_i^p\} = f_{k-\text{means}}(\{\boldsymbol{x}_{\boldsymbol{\theta}, i}^t\}, K), \tag{24}
\end{equation}
where $K$ is the preset number of clusters. Therefore, we can calculate the differential embeddings of positive samples and negative samples according to the target-domain pseudo-labels:
\begin{equation}
    \boldsymbol{x}_{\boldsymbol{\theta}, i,j}^{t,+} = \boldsymbol{x}_{\boldsymbol{\theta}, i}^t - \boldsymbol{x}_{\boldsymbol{\theta}, j}^t, \quad y_i^p = y_j^p, \tag{25}
\end{equation}
\begin{equation}
    \boldsymbol{x}_{\boldsymbol{\theta}, i,j}^{t,-} = \boldsymbol{x}_{\boldsymbol{\theta}, i}^t - \boldsymbol{x}_{\boldsymbol{\theta}, j}^t, \quad y_i^p \neq y_j^p. \tag{26}
\end{equation}
For ease of writing, we use $m = 0, \ldots, N_1^t - 1$ and $l = 0, \ldots, N_0^t - 1$ to re-index target domain positive set $X_1^t$ and negative samples $X_0^t$. For MG-GLRTML, we can update the statistical distribution parameters according to the following formula:
\begin{equation}
    \hat{\boldsymbol{\Sigma}}_1 = \frac{1}{N} \sum_{m=0}^{N_1^t - 1} \left( \boldsymbol{x}_{\boldsymbol{\theta}, m}^{t,+} \right)^\top \boldsymbol{x}_{\boldsymbol{\theta}, m}^{t,+}, \quad \boldsymbol{x}_{\boldsymbol{\theta}, l}^{t,-} \in X_1^t, \tag{27}
\end{equation}
\begin{equation}
    \hat{\boldsymbol{\Sigma}}_0 = \frac{1}{N} \sum_{l=0}^{N_0^t - 1} \left( \boldsymbol{x}_{\boldsymbol{\theta}, l}^{t,-} \right)^\top \boldsymbol{x}_{\boldsymbol{\theta}, l}^{t,-}, \quad \boldsymbol{x}_{\boldsymbol{\theta}, l}^{t,-} \in X_0^t. \tag{28}
\end{equation}
For GMM-GLRTML, we can use EM algorithm \cite{ref58} to update the estimated distribution parameters $\boldsymbol{\vartheta}_1$ and $\boldsymbol{\vartheta}_0$. The update process can refer to Eq.~(21) and Eq.~(22). The only difference is that the embeddings of the positive and negative samples are calculated based on Eq.~(25) and Eq.~(26). Then calculate the estimated similarity score based on the updated statistical distribution parameters $\boldsymbol{\vartheta}_1^t$ and $\boldsymbol{\vartheta}_0^t$:
\begin{equation}
    \hat{s}(i, j) = \log \left( \frac{\hat{p}(\boldsymbol{x}_{\boldsymbol{\theta}, ij} | \boldsymbol{\vartheta}_1^t)}{\hat{p}(\boldsymbol{x}_{\boldsymbol{\theta}, ij} | \boldsymbol{\vartheta}_0^t)} \right). \tag{29}
\end{equation}
Finally, estimated similarity scores are sorted in descending order. In the above way, we obtain pseudo-labels from the target domain and update the metric space by updating statistical parameters instead of updating neural network parameters, thereby achieving UDA CBRSOR. It is worth mentioning that for MG-GLRTML, CPLFPA only needs to update $d^2$ parameters in one step. For GMM-GLRTML, only $\frac{(K_0 + K_1)d(d + 2)}{2}$ parameters need to be updated, and the fast convergence characteristics of K-means \cite{ref60} also ensure the high efficiency of CPLFPA. Algorithm 3 gives the specific process.

\section{Experiments}

In this section, we evaluate the performance of GLRTML and CPLFPA on remote sensing images with ship and aircraft scenes. First, we present the experimental setup, including detailed descriptions of the datasets used, the evaluation metrics, and the implementation details of our methods. Second, we demonstrate the effectiveness of GLRTML and CPLFPA through extensive ablation experiments. Third, we perform further analyses to examine the impact of the statistical model on GLRTML, investigate the robustness of CPLFPA to different preset numbers of clusters, assess the adaptability of the algorithm to different neural network architectures, and evaluate the computational efficiency of CPLFPA. Finally, we compare our methods with other state-of-the-art approaches to demonstrate the superior performance of our proposed method.

\begin{algorithm}[H]
\caption{CPLFPA}
\label{alg:cplfpa}

\textbf{Input:} \\
Target-domain images set $\{I_i^t\}$ and corresponding label set. Neural network $\phi$ with pre-trained parameters $\tilde{\boldsymbol{\theta}}$ from GLRTML algorithm. The preset number of clusters $K$. \\

\textbf{Update parameters:}
\begin{algorithmic}[1]
    \For{each target-domain image $I_i^t$ }
        \State Forward propagations $\boldsymbol{x}_{\boldsymbol{\theta}, i}^t = \phi(I_i^t, \tilde{\boldsymbol{\theta}})$.
    \EndFor
    \State Calculate pseudo-labels based on K-means algorithm \cite{ref60}: $\{y_i^p\} = f_{\text{k-means}}(\{\boldsymbol{x}_{\boldsymbol{\theta}, i}^t\}, K)$.
    \State Calculate differential embeddings.
    \State Calculate updated statistical distribution parameters $\boldsymbol{\vartheta}_1^t$ and $\boldsymbol{\vartheta}_0^t$ based on Eq.~(27) and Eq.~(28) for MG-GLRTML, or based on EM algorithm \cite{ref58} for GMM-GLRTML.
\end{algorithmic}

\textbf{Test:}

\textbf{Input:} \\
Images set $\{I_i\}_q$ and $\{I_j\}_g$ from query set and gallery set. Trained neural network $\phi$ with pre-trained parameters $\tilde{\boldsymbol{\theta}}$, and updated statistical distribution parameters $\boldsymbol{\vartheta}_1^t$ and $\boldsymbol{\vartheta}_0^t$. \\

\begin{algorithmic}[1]
    \For{each image $I_i$ in query set}
        \State Forward propagations $\boldsymbol{x}_{\boldsymbol{\theta}_{T-1, B-1}, i} = \phi(I_i, \boldsymbol{\theta}_{T-1, B-1})$.
        \For{each image $I_j$ in gallery set \textbf{do}}
            \State Forward propagations $\boldsymbol{x}_{\boldsymbol{\theta}_{T-1, B-1}, j} = \phi(I_j, \boldsymbol{\theta}_{T-1, B-1})$.
            \State Calculate differential embedding $\boldsymbol{x}_{\boldsymbol{\theta}_{T-1, B-1}, ij}$.
            \State Calculate estimated similarity score based on Eq.~(29).
        \EndFor
        \State Sort estimated similarity score in descending order to get the retrieval results for $I_i$.
    \EndFor
\end{algorithmic}
\end{algorithm}

\subsection{Experimental Settings}

\begin{table*}[h]
\footnotesize 
\centering
\caption{Ablation Experiment Results on CBRSOR-FGSRSI} 
\label{table:I} 
\begin{tabular}{ccccccc}
\toprule
\textbf{Backbone} & \textbf{GLRTML-TR} & \textbf{GLRTML-TE} & \textbf{CPLFPA-17} & \textbf{mAP (\%)} & \textbf{R@50 (\%)} & \textbf{P@50 (\%)} \\ 
\midrule
Resnet-50         &                    &                    &                    & 77.6              & 73.9              & 58.6              \\ 
Resnet-50         & \checkmark         &                    &                    & 80.2              & 75.8              & 60.3              \\ 
Resnet-50         & \checkmark         & \checkmark         &                    & \textbf{81.1}     & 76.6              & \textbf{60.9}     \\ 
Resnet-50         & \checkmark         & \checkmark         & \checkmark         & 80.9              & \textbf{76.8}     & \textbf{60.9}     \\ 
\bottomrule
\end{tabular}
\end{table*}

\begin{table*}[h]
\footnotesize
\centering
\caption{Ablation Experiment Results on CBRSOR-MAR}
\label{table:II}
\begin{tabular}{ccccccc}
\toprule
\textbf{Backbone} & \textbf{GLRTML-TR} & \textbf{GLRTML-TE} & \textbf{CPLFPA-21} & \textbf{mAP (\%)} & \textbf{R@50 (\%)} & \textbf{P@50 (\%)} \\ 
\midrule
Resnet-50         &                    &                    &                    & 80.7              & 59.4              & 76.9              \\ 
Resnet-50         & \checkmark         &                    &                    & 81.9              & 60.0              & 77.5              \\ 
Resnet-50         & \checkmark         & \checkmark         &                    & 82.1              & \textbf{60.3}     & \textbf{77.7}     \\ 
Resnet-50         & \checkmark         & \checkmark         & \checkmark         & \textbf{82.4}     & 60.0              & 77.5              \\ 
\bottomrule
\end{tabular}
\end{table*}

\begin{table*}[h]
\footnotesize
\centering
\caption{Ablation Experiment Results on UDA-CBRSOR-FGSRSI}
\label{table:III}
\begin{tabular}{ccccccc}
\toprule
\textbf{Backbone} & \textbf{GLRTML-TR} & \textbf{GLRTML-TE} & \textbf{CPLFPA-8} & \textbf{mAP (\%)} & \textbf{R@50 (\%)} & \textbf{P@50 (\%)} \\ 
\midrule
Resnet-50         &                    &                    &                    & 51.2              & 31.9              & 60.3              \\ 
Resnet-50         & \checkmark         &                    &                    & 48.9              & 31.7              & 59.0              \\ 
Resnet-50         & \checkmark         & \checkmark         &                    & 44.9              & 28.1              & 56.5              \\ 
Resnet-50         & \checkmark         & \checkmark         & \checkmark         & \textbf{53.3}     & \textbf{33.3}     & \textbf{61.0}     \\ 
\bottomrule
\end{tabular}
\end{table*}

\begin{table*}[h]
\footnotesize
\centering
\caption{Ablation Experiment Results on UDA-CBRSOR-MAR}
\label{table:IV}
\begin{tabular}{ccccccc}
\toprule
\textbf{Backbone} & \textbf{GLRTML-TR} & \textbf{GLRTML-TE} & \textbf{CPLFPA-10} & \textbf{mAP (\%)} & \textbf{R@50 (\%)} & \textbf{P@50 (\%)} \\ 
\midrule
Resnet-50         &                    &                    &                    & 65.4              & 10.4              & 77.3              \\ 
Resnet-50         & \checkmark         &                    &                    & 67.9              & 10.3              & 78.3              \\ 
Resnet-50         & \checkmark         & \checkmark         &                    & 66.1              & 10.3              & 77.7              \\ 
Resnet-50         & \checkmark         & \checkmark         & \checkmark         & \textbf{69.4}     & \textbf{10.6}     & \textbf{78.8}     \\ 
\bottomrule
\end{tabular}
\end{table*}

\subsubsection*{\bf 1) Dataset Description and Evaluation Metric}

To validate the performance of our proposed methods on the CBRSOR task, we reorganize two remote sensing fine-grained recognition datasets—FGSRSI-23 \cite{ref1} and MAR20 \cite{ref61}—to meet the specific requirements of CBRSOR. The details of these datasets are as follows:

FGSRSI-23 \cite{ref1} is a fine-grained ship recognition dataset published by Xiong et al. It consists of 1,067 images in 23 fine-grained ship subcategories. The images are primarily collected from Google Earth and the SuperView-1 satellite. An advantage of FGSRSI-23 is that each image contains only one ship target, making it particularly suitable for CBRSOR tasks. To minimize randomness in the experimental results, we remove categories with less than 10 samples, resulting in a processed dataset of 16 categories and 1,038 images. Following the CBRSOR experimental settings, we divide the dataset into training, query, and gallery sets. For the conventional CBRSOR experimental setting, the training set contains 516 ship samples, the query set contains 202 samples, and the gallery set contains 320 samples, with all sets including all ship categories. To simulate more realistic CBRSOR scenarios, we add 2,065 background images to the gallery set. This reorganized dataset is called CBRSOR-FGSRSI. For the unsupervised domain adaptation CBRSOR experimental setting, we partition the dataset into 9 categories for training and 7 categories for testing. The training, query, and gallery sets contain 545, 96, and 397 ship samples, respectively. In addition, we add 2,065 background images to the gallery set. This dataset is called UDA-CBRSOR-FGSRSI.

MAR20 \cite{ref61} is a fine-grained aircraft detection and recognition dataset released by Yu et al., containing 22,341 instances across 20 fine-grained aircraft subcategories. The images are primarily collected from Google Earth. To facilitate experiments, we crop the aircraft instances according to the detection labels provided by MAR20 and randomly select 9,697 instances. We then divide the dataset into training, query, and gallery sets. For the conventional CBRSOR experimental setting, the training set contains 7,870 aircraft instances, the query set contains 633 instances, and the gallery set contains 1,194 instances, with all sets encompassing all aircraft categories. To simulate more realistic CBRSOR scenarios, we add 464 background images to the gallery set. This reorganized dataset is referred to as CBRSOR-MAR. For the unsupervised domain adaptation CBRSOR experimental setting, we partition the dataset into 12 categories for training and 8 categories for testing. The training, query, and gallery sets contain 6,103, 233, and 3,361 aircraft instances, respectively. Additionally, we add 464 remote sensing background images to the gallery set. This dataset is designated as UDA-CBRSOR-MAR.

To comprehensively evaluate the performance of our methods, we employ a set of performance metrics, including mean Average Precision (mAP), Recall at top-50 retrieval results (R@50), and Precision at top-50 retrieval results (P@50).

\subsubsection*{\bf 2) Implementation Details}

We reproduce all methods reported in this article and ensure fair backbone network and consistent hyperparameters. In terms of the backbone network, we use the popular convolution-based neural network Resnet-50 \cite{ref62} and the Transformer-based neural network Swin Transformer V2 (Swin-TV2) \cite{ref63} as embedding extractors. The hyperparameters used in this article are reported as follows. The batch size $M$ is set to 64, and the embedding length $d$ is 64. The epoch of first training stage $T_0$ is 100 and second training stage $T_1 - T_0$ is 50. The momentum of SGD is 0.9, the learning rates of first training stage and second training stage are 0.05 and 0.005, respectively. In the loss function, the temperature coefficient $\nu$ is 0.001, weighting coefficient $\alpha$ is 1, and the weight decay coefficient is 0.0005.

Next, we introduce experimental environment in detail. For a fair comparison, all the experiments are conducted in Python version 3.8.18 and PyTorch version 2.1.2, with Intel(R) Xeon(R) Platinum 8358P CPU and NVIDIA A800-80GB GPU. The system is deployed on a computer running the Ubuntu 20.04 operating system, and CUDA toolkit version is 11.4.

\subsection{Ablation Experiment}

We perform detailed ablation experiments to verify the effectiveness of GLRTML and CPLFPA. We mainly discuss MG-GLRTML in the ablation experiment, and the performance analysis of GMM-GLRTML is given in analytical experiments. 

Table I and Table II show the ablation experiment results for the CBRSOR task of our algorithm on FGSRSI-23 and MAR20, respectively. GLRTML-TR and GLRTML-TE represent the use of statistical models in the training and testing phases, respectively. If the use of statistical models is dropped during the testing phase, cosine similarity is used to compute the similarity between embeddings. CPLFPA- means CPLFPA with a given number of clusters. By adding GLRTML-TR to the baseline, significant performance improvements are achieved even when cosine similarity is used as the metric space during the testing phase. This shows that GLRTML introduces statistical information of the global data during the training phase, which avoids overfitting of the neural network and improves the feature expression ability of the neural network, this result supports Discussion 2. By introducing a statistical model during the testing stage, the performance is further significantly improved, which is intuitive because the metric space of the training and testing processes remains consistent under this condition, reflecting the superiority of the metric space derived based on the likelihood ratio. However, Table I and Table II do not show the effectiveness of CPLFPA. This is because the training and test sets are identically distributed, and there is no need to adjust the statistical parameters.

Table III and Table IV show the experimental results of our algorithm in the unsupervised domain adaptation CBRSOR experimental setting on FGSRSI-23 and MAR20, respectively. Without the use of CPLFPA, GLRTML does not bring significant performance gains and even leads to performance degradation. This is because the training and test sets are not identically distributed under this condition, and the performance of GLRTML is sensitive to the estimation of statistical parameters. CPLFPA brings significant performance improvements, illustrating the feasibility and effectiveness of adjusting the statistical parameters of GLRTML based on pseudo-labels to implement UDA CBRSOR.

\begin{table*}[h]
\footnotesize 
\centering
\caption{The Impact of Different Preset Number of Clusters on Performance} 
\label{table:V} 
\begin{tabular}{cc|cc|cc|cc}
\toprule
\multicolumn{2}{c|}{\textbf{CBRSOR-FGSRSI}} & \multicolumn{2}{c|}{\textbf{CBRSOR-MAR}} & \multicolumn{2}{c|}{\textbf{UDA-CBRSOR-FGSRSI}} & \multicolumn{2}{c}{\textbf{UDA-CBRSOR-MAR}} \\ \midrule
\textbf{\textit{K}} & \textbf{mAP (\%)} & \textbf{\textit{K}} & \textbf{mAP (\%)} & \textbf{\textit{K}} & \textbf{mAP (\%)} & \textbf{\textit{K}} & \textbf{mAP (\%)} \\ \midrule
14 & 80.1 & 18 & 80.1 & 5 & 49.0 & 6 & 69.7 \\
15 & 80.7 & 19 & 81.7 & 6 & 53.2 & 7 & 69.0 \\
16 & 80.2 & 20 & 82.1 & 7 & 53.4 & \textbf{8} & \textbf{70.1} \\
\underline{17} & \underline{80.9} & \underline{21} & \underline{82.4} & \underline{8} & \underline{53.3} & \underline{9} & \underline{69.4} \\
18 & 80.7 & 22 & 82.6 & 9 & 54.5 & 10 & 69.4 \\
\textbf{19} & \textbf{81.0} & \textbf{23} & \textbf{82.7} & 10 & 57.0 & 11 & 68.9 \\
\textbf{20} & \textbf{81.0} & \textbf{24} & \textbf{82.7} & \textbf{11} & \textbf{58.1} & 12 & 69.3 \\
\bottomrule
\end{tabular}
\end{table*}

\begin{table*}[h]
\centering
\caption{Ablation Experiment Results on CBRSOR-FGSRSI Using Swin Transformer V2}
\label{table:VI}
\begin{tabular}{ccccccc}
\toprule
\textbf{Backbone} & \textbf{GLRTML-TR} & \textbf{GLRTML-TE} & \textbf{CPLFPA-17} & \textbf{mAP (\%)} & \textbf{R@50 (\%)} & \textbf{P@50 (\%)} \\ 
\midrule
Swin-TV2 &                    &                    &                    & 80.7              & 76.7              & 59.1              \\ 
Swin-TV2 & \checkmark         &                    &                    & 81.8              & \textbf{78.2}     & 60.5              \\ 
Swin-TV2 & \checkmark         & \checkmark         &                    & 81.9              & 77.9              & 60.8              \\ 
Swin-TV2 & \checkmark         & \checkmark         & \checkmark         & \textbf{82.1}     & \textbf{78.2}     & \textbf{60.9}     \\ 
\bottomrule
\end{tabular}
\end{table*}

\begin{table*}[h]
\centering
\caption{Ablation Experiment Results on UDA-CBRSOR-FGSRSI Using Swin Transformer V2}
\label{table:VII}
\begin{tabular}{ccccccc}
\toprule
\textbf{Backbone} & \textbf{GLRTML-TR} & \textbf{GLRTML-TE} & \textbf{CPLFPA-8} & \textbf{mAP (\%)} & \textbf{R@50 (\%)} & \textbf{P@50 (\%)} \\ 
\midrule
Swin-TV2 &                    &                    &                    & 63.4              & 31.9              & 66.1              \\ 
Swin-TV2 & \checkmark         &                    &                    & 67.9              & \textbf{36.9}     & 68.8              \\ 
Swin-TV2 & \checkmark         & \checkmark         &                    & 63.4              & 32.6              & 66.5              \\ 
Swin-TV2 & \checkmark         & \checkmark         & \checkmark         & \textbf{71.6}     & 36.2     & \textbf{69.0}     \\ 
\bottomrule
\end{tabular}
\end{table*}

\begin{table*}[h]
\centering
\caption{The Running Time of CPLFPA on UDA-CBRSOR-FGSRSI on CPU Platform}
\label{table:VIII}
\begin{tabular}{cc|ccc|cc}
\toprule
\textbf{Number of} & \textbf{Number of} & \textbf{Clustering} & \textbf{Calculating} & \textbf{Updating} & \textbf{Total} & \textbf{mAP} \\
\textbf{positive samples} & \textbf{negative samples} & \textbf{time (s)} & \textbf{differential embeddings} & \textbf{parameters} & \textbf{Time (s)} & \textbf{(\%)} \\
& & & \textbf{time (s)} & \textbf{time (s)} & & \\
\midrule
20,000 & 20,000 & 2.4 & 1.4 & 2.5 & 6.3 & 53.3 \\ 
30,000 & 30,000 & 2.4 & 1.7 & 3.6 & 7.7 & 55.3 \\ 
40,000 & 40,000 & 2.4 & 2.1 & 3.9 & 8.4 & 55.2 \\ 
\bottomrule
\end{tabular}
\end{table*}

\subsection{Analytical Experiments}

\begin{figure}[!t]
\centering
\includegraphics[width=3.5in]{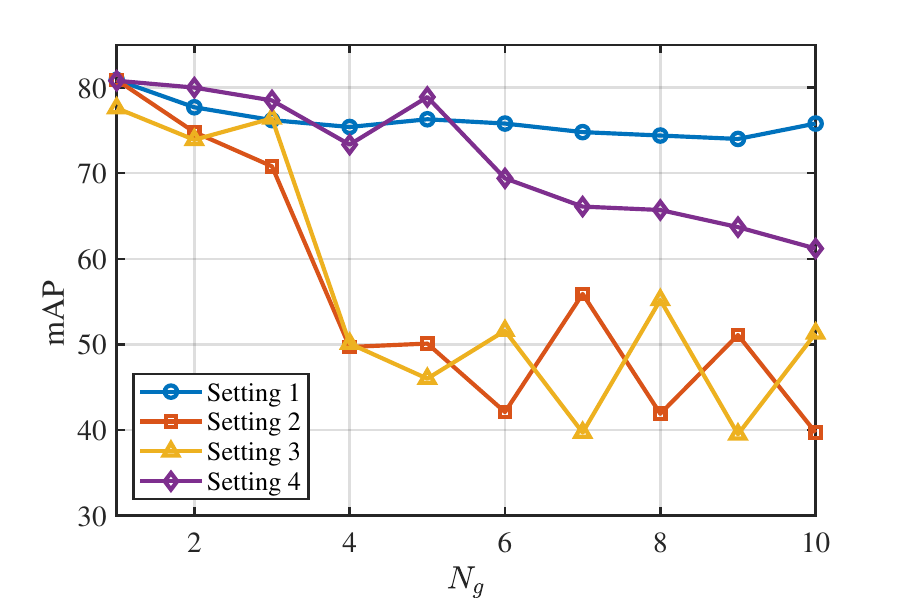}
\caption{Performance of the proposed method under different number of Gaussian component settings on CBRSOR-FGSRSI. Setting 1: we set $K_0 = N_g$, $K_1 = 1$ in GLRTML, and use CPLFPA. Setting 2: we set $K_0 = N_g$, $K_1 = N_g$ in GLRTML, and use CPLFPA. Setting 3: we set $K_0 = N_g$, $K_1 = 1$ in GLRTML, and cancel CPLFPA. Setting 4: we set $K_0 = N_g$, $K_1 = N_g$ in GLRTML, constrain the covariance matrix of each Gaussian component of the GMM to be a diagonal matrix, and use CPLFPA.}
\label{fig_4}
\end{figure}

\subsubsection*{\bf 1) Impact of Statistical Model Selection on Algorithm Performance}
We design GMM-GLRTML and explore the impact of the statistical model on performance by adjusting the number of Gaussian components of hypothesis $H_0$ and hypothesis $H_1$, i.e., $K_0$ and $K_1$. The basis for designing experiments in this way is that when there are enough Gaussian components, GMM theoretically has the ability to fit any distribution \cite{ref58}, and when $K_0=1$ and $K_1=1$, GMM-GLRTML degenerates into MG-GLRTML, Fig.~4 gives the experimental results.

We observe that as the number of Gaussian components increases, different settings consistently show performance degradation. In addition, the comparison between setting 1 and setting 2 shows that when $K_0$ and $K_1$ increase at the same time, the performance degradation is further aggravated than setting $K_1=1$. A comparison between setting 2 and setting 4 shows that constraining the covariance of each Gaussian component to a diagonal matrix can significantly alleviate the deterioration of performance. These all show that although the increase in model complexity can theoretically fit any distribution, due to the limited data, it may lead to overfitting and degrade performance.

To further validate our hypothesis that overfitting causes the performance of GLRTML to decrease as the complexity of the statistical model increases, we visualize the embedding distributions and fitting performances of MG-GLRTML and GMM-GLRTML with higher complexity settings on the training and test sets of the CBRSOR-FGSRSI dataset in Fig.~5. Comparing Fig.~5(a) and Fig.~5(c), we observe that GMM-GLRTML fits the training set embeddings better than MG-GLRTML, indicating that the higher complexity model captures the training data more accurately. However, examining the test set results in Fig.~5(b) and Fig.~5(d), we find that MG-GLRTML successfully learns the desired pattern: the feature embeddings of positive sample pairs are concentrated near the center with surrounding dispersion, and the embeddings of negative sample pairs exhibit a sparse central distribution. In contrast, GMM-GLRTML, despite its higher parameter complexity, significantly overfits and fails to learn the correct pattern on the test set. This overfitting leads to a significant drop in performance when generalizing to unseen data. Therefore, considering both the performance and the computational efficiency of the algorithm, we recommend using MG-GLRTML instead of GMM-GLRTML.

In addition, we try to disable CPLFPA and compare setting 1 and setting 3, then we find another interesting conclusion: when the model complexity is high, even though the training and test sets are identically distributed, using CPLFPA can still improve the algorithm performance. We speculate that this is because CPLFPA generates pseudo-labels based on the distribution of the embeddings and re-estimates the distribution parameters. The results of pseudo-labels are relatively smooth, so it is not easy to overfit to new distributions based on pseudo-labels.

\begin{figure*}[!t]
\centering

\begin{minipage}[t]{0.4\textwidth}
    \centering
    \subfloat[]{\includegraphics[width=\linewidth]{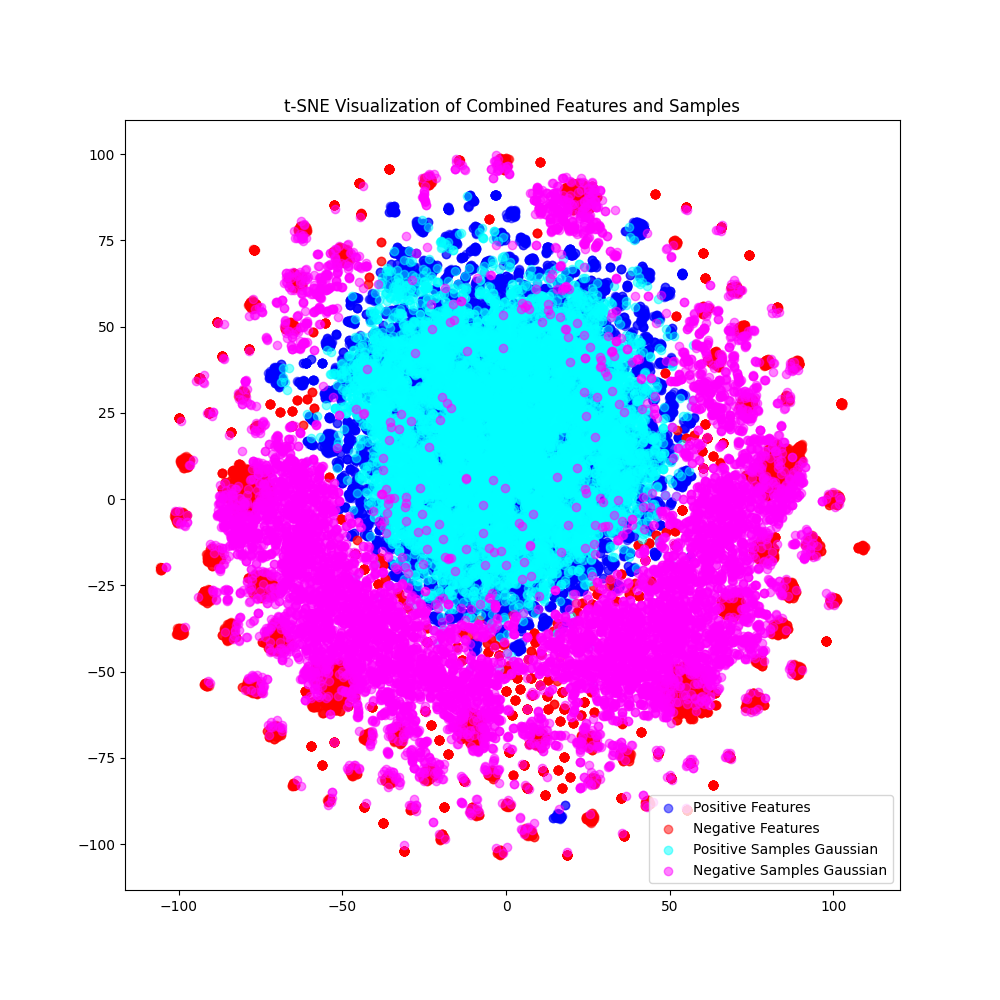}%
    \label{fig5_first_case}}
\end{minipage}
\hspace{0.04\textwidth}
\begin{minipage}[t]{0.4\textwidth}
    \centering
    \subfloat[]{\includegraphics[width=\linewidth]{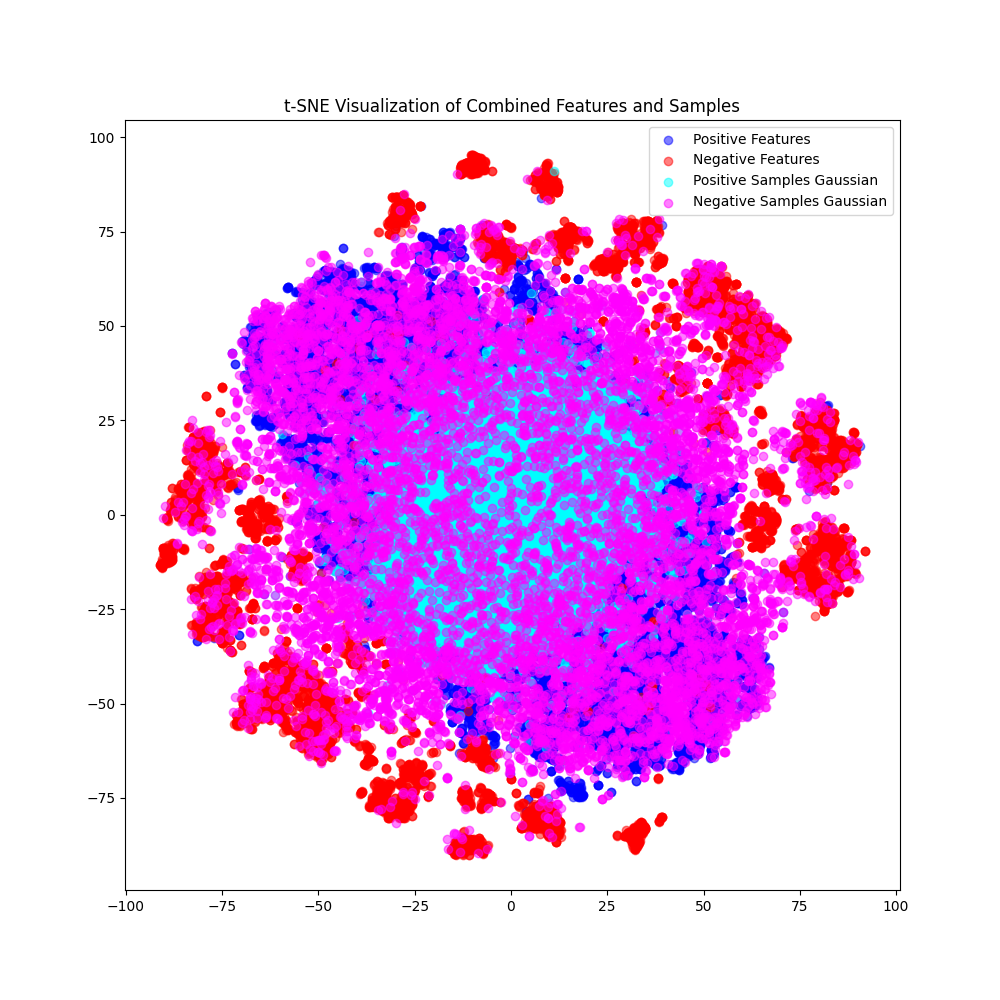}%
    \label{fig5_second_case}}
\end{minipage}

\vspace{0.04\textwidth} 
\begin{minipage}[t]{0.4\textwidth}
    \centering
    \subfloat[]{\includegraphics[width=\linewidth]{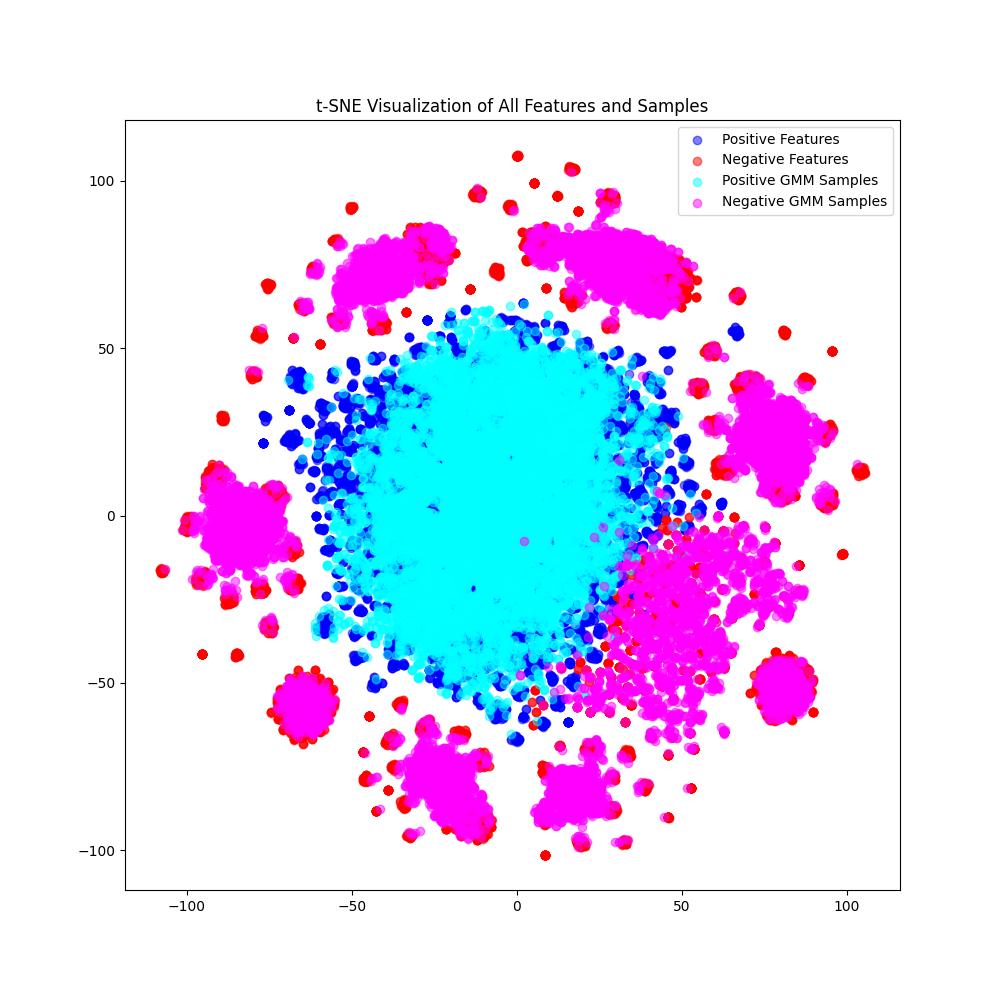}%
    \label{fig5_third_case}}
\end{minipage}
\hspace{0.04\textwidth} 
\begin{minipage}[t]{0.4\textwidth}
    \centering
    \subfloat[]{\includegraphics[width=\linewidth]{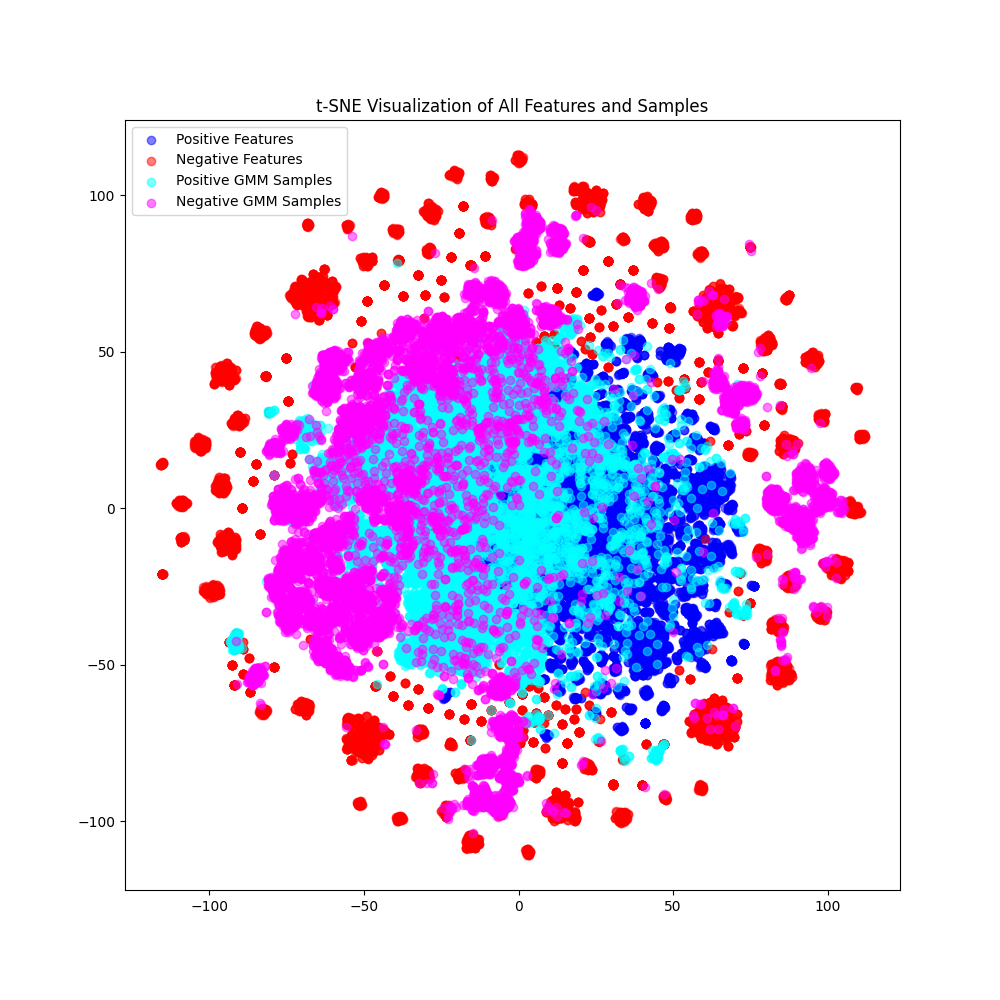}%
    \label{fig5_fourth_case}}
\end{minipage}

\caption{Visualization results of the statistical model fitting capabilities of MG-GLRTML and GMM-GLRTML ($K_1 = 1$, $K_0 = 10$). (a) MG-GLRTML visualization results on the training set. (b) MG-GLRTML visualization results on the test set. (c) GMM-GLRTML visualization results on the training set. (d) GMM-GLRTML visualization results on the test set. The blue points in the figure represent the embeddings of the real positive samples, visualized using t-SNE \cite{ref64}. The red points represent real negative sample embeddings, visualized using t-SNE \cite{ref64}. The cyan points represent the fitted embeddings of positive samples by the statistical model, which are obtained through Monte Carlo simulation \cite{ref65} based on the statistical model and then visualized using t-SNE \cite{ref64}. The pink points represent the fitted embeddings of the negative samples by the statistical model, which are obtained through Monte Carlo simulation \cite{ref65} based on the statistical model and then visualized using t-SNE \cite{ref64}.}
\label{fig_5}
\end{figure*}

\subsubsection*{\bf 2) Robustness of CPLFPA to Preset Number of Clusters}

  For the following two reasons, we believe it is necessary to discuss the impact of preset number of clusters   on CPLFPA. (1) K-means must be performed in CPLFA to obtain pseudo-labels, and the preset number of clusters will have a certain impact on the pseudo-label results. (2) The preset number of clusters may be closely related to the number of categories in the target domain, but the real number of categories in the target domain may not be available. Table V shows the experimental results. We set the sum of the number of real categories and the background in the target domain as the base number of clusters (the experimental group corresponding to the underlined in Table V), and a control experimental group in which the base number of clusters was increased or decreased by 1 to 3. We find that the base number of clusters may not necessarily achieve the best performance. At the same time, we find that slightly increasing the number of clusters does not reduce performance, and even slightly improves performance. This may be because background images can also be divided into several categories, and their embeddings do not necessarily cluster together. Increasing the number of clusters can yield pseudo-labels that are more consistent with the semantic distribution. On the other hand, we find that reducing the number of clusters hurts performance in most cases, because reducing the number of clusters prevents the algorithm from correctly assigning pseudo-labels to each object, forcing different categories to merge in the feature space, and causing confusion. In summary, we recommend setting a reasonably large number of clusters when the number of categories in the target domain is unknown, and we believe that within a reasonable range, CPLFPA is robust to changes in the number of clusters.

\subsubsection*{\bf 3) Robustness of Method to Backbone}

Previous experiments use the classical convolutional neural network ResNet-50 \cite{ref62} as a backbone. However, the proposed GLRTML and CPLFPA are independent of the neural network structure. To verify the universality of our algorithm, we use the advanced transformer-based backbone network Swin Transformer V2 (Swin-TV2) \cite{ref63} to perform the performance verification. Table VI and Table VII show the experimental results on CBRSOR-FGSRSI and UDA-CBRSOR-FGSRSI, respectively. After replacing the backbone network, the conclusion on the effectiveness of the algorithm is consistent with the previous analysis: For conventional CBRSOR tasks, the introduction of statistical models in both the training and testing phases can effectively improve the performance. For UDA CBRSOR tasks, CPLFPA is very effective.

\subsubsection*{\bf 4) The Efficiency of CPLFPA}

The most prominent advantage of CPLFPA over the existing UDA method is its excellent efficiency. Table VIII shows the running time of the CPLFPA algorithm. CPLFPA is implemented based on GM-GLRTML. The target domain of UDA-CBRSOR-FGSRSI has a total of 2558 images, and the number of clusters is set to 8. The number of positive and negative samples can affect the runtime and performance. Therefore, we set different numbers of positive and negative samples for experimental analysis. To emulate edge computing platforms, our CPLFPA runs entirely on the CPU. As shown in Table VIII, our algorithm is extremely fast and maintains a running time within 10s under different settings of the number of positive and negative samples. This is much faster than the existing UDA method and is expected to be truly implemented as a real-time UDA method on edge computing platforms. In addition, we also find that appropriately increasing the number of positive and negative samples can help improve the performance of CPLFPA, because adapting to appropriately larger number of samples helps to correctly estimate the statistical model parameters.

\subsection{Comparison with State-of-the-Art Methods}

\begin{table}[h]
\centering
\caption{Comparison with State-of-the-Art Methods on CBRSOR-FGSRSI}
\label{table:IX}
\begin{tabular}{lccc}
\toprule
\textbf{Method} & \textbf{mAP (\%)} & \textbf{R@50 (\%)} & \textbf{P@50 (\%)} \\
\midrule
Identity loss \cite{ref33}       & 77.3              & 70.6              & 57.5              \\
Sphere loss \cite{ref34}         & 75.4              & 72.0              & 57.7              \\
Triplet loss \cite{ref37}        & 76.4              & 72.3              & 57.7              \\
Lifted loss \cite{ref38}         & 79.4              & 76.0              & 60.4              \\
Instance loss \cite{ref39}       & 75.1              & 71.5              & 58.1              \\
Contrastive loss \cite{ref36}    & 78.1              & 74.2              & 59.3              \\
Circle loss \cite{ref41}         & 73.3              & 67.4              & 57.1              \\
SP loss \cite{ref40}             & 77.6              & 74.6              & 58.7              \\
UCE loss \cite{ref35}            & \underline{80.9}  & 76.4              & 60.0              \\
CCL loss \cite{ref22}            & 75.2              & 72.1              & 56.4              \\
\textbf{Ours (K=17)}             & \underline{80.9}  & \underline{76.8}  & \underline{60.9}  \\
\textbf{Ours (K=20)}             & \textbf{81.0}     & \textbf{77.3}     & \textbf{61.0}     \\
\bottomrule
\end{tabular}
\end{table}

\begin{table}[h]
\centering
\caption{Comparison with State-of-the-Art Methods on CBRSOR-MAR}
\label{table:X}
\begin{tabular}{lccc}
\toprule
\textbf{Method} & \textbf{mAP (\%)} & \textbf{R@50 (\%)} & \textbf{P@50 (\%)} \\
\midrule
Identity loss \cite{ref33}       & 81.6              & \underline{60.0}   & 77.2              \\
Sphere loss \cite{ref34}         & 74.5              & 54.0               & 72.3              \\
Triplet loss \cite{ref37}        & 78.5              & 59.2               & 76.8              \\
Lifted loss \cite{ref38}         & 79.5              & 58.9               & 76.2              \\
Instance loss \cite{ref39}       & 78.3              & 57.3               & 75.3              \\
Contrastive loss \cite{ref36}    & 78.7              & 57.0               & 76.0              \\
Circle loss \cite{ref41}         &81.9               & 59.4               & \textbf{78.3}     \\
SP loss \cite{ref40}             & 81.1              & 59.8               & 76.9              \\
UCE loss \cite{ref35}            & 80.2              & 59.5               & 76.1              \\
CCL loss \cite{ref22}            & 75.2              & 60.3               & 76.5              \\
\textbf{Ours (K=21)}             & \underline{82.4}  & \underline{60.0}   & 77.5              \\
\textbf{Ours (K=24)}             & \textbf{82.7}     & \textbf{60.1}      & \underline{77.6}  \\
\bottomrule
\end{tabular}
\end{table}

\begin{table}[h]
\centering
\caption{Comparison with State-of-the-Art Methods on UDA-CBRSOR-FGSRSI}
\label{table:XI}
\begin{tabular}{lccc}
\toprule
\textbf{Method} & \textbf{mAP (\%)} & \textbf{R@50 (\%)} & \textbf{P@50 (\%)} \\
\midrule
Identity loss \cite{ref33} & 51.2 & 31.9 & 60.3 \\
Sphere loss \cite{ref34} & 49.1 & 32.3 & \underline{61.3} \\
Triplet loss \cite{ref37} & 31.4 & 22.3 & 44.3 \\
Lifted loss \cite{ref38} & 47.2 & 27.4 & 56.1 \\
Instance loss \cite{ref39} & 47.5 & 29.7 & 58.7 \\
Contrastive loss \cite{ref36} & 48.4 & 26.0 & 55.8 \\
Circle loss \cite{ref41} & \underline{53.6} & 27.8 & 60.7 \\
SP loss \cite{ref40} & 47.4 & 29.9 & 58.2 \\
UCE loss \cite{ref35} & 44.7 & 29.8 & 54.9 \\
CCL loss \cite{ref22} & 48.0 & 28.1 & 57.5 \\
\textbf{Ours (K=8)} & 53.3 & \textbf{33.3} & 61.0 \\
\textbf{Ours (K=11)} & \textbf{58.1} & \underline{33.0} & \textbf{62.3} \\
\bottomrule
\end{tabular}
\end{table}

\begin{table}[h]
\centering
\caption{Comparison with State-of-the-Art Methods on UDA-CBRSOR-MAR}
\label{table:XII}
\begin{tabular}{lccc}
\toprule
\textbf{Method} & \textbf{mAP (\%)} & \textbf{R@50 (\%)} & \textbf{P@50 (\%)} \\
\midrule
Identity loss \cite{ref33} & 65.4 & 10.4 & 77.3 \\
Sphere loss \cite{ref34} & 67.4 & 10.0 & 76.2 \\
Triplet loss \cite{ref37} & 60.7 & 9.3 & 70.7 \\
Lifted loss \cite{ref38} & 60.6 & 9.8 & 73.2 \\
Instance loss \cite{ref39} & 67.4 & 10.3 & 77.8 \\
Contrastive loss \cite{ref36} & 63.1 & 9.8 & 73.7 \\
Circle loss \cite{ref41} & 66.9 & 10.4 & 78.5 \\
SP loss \cite{ref40} & \textbf{69.4} & 10.4 & \textbf{79.4} \\
UCE loss \cite{ref35} & 62.4 & 10.0 & 74.8 \\
CCL loss \cite{ref22} & \underline{68.0} & \underline{10.5} & \underline{79.3} \\
\textbf{Ours (K=9)} & \textbf{69.4} & 10.0 & 77.0 \\
\textbf{Ours (K=10)} & \textbf{69.4} & \textbf{10.6} & 78.8 \\
\bottomrule
\end{tabular}
\end{table}

Tables IX, X, XI, and XII present quantitative performance comparisons between our proposed method and existing SOTA methods on the CBRSOR-FGSRSI, CBRSOR-MAR, UDA-CBRSOR-FGSRSI, and UDA-CBRSOR-MAR datasets, respectively. In these experiments, all methods utilize ResNet-50 as the default backbone. Our method employs GLRTML and CPLFPA by default, with the number of clusters in CPLFPA set to the base cluster number. Although this setting may not yield optimal performance for our method, we maintain it to ensure fairness in the experimental protocol and avoid tuning hyperparameters specific to our approach. For completeness, we also report the performance of our method with a larger number of clusters. Additional visual examples of the top-10 retrieval results for our method and representative methods are provided in Appendix B. The experimental results demonstrate that our method consistently achieves superior performance across different datasets, reaching SOTA levels.

\section{Conclusion}
In this article, we address the problem of content-based remote sensing object retrieval (CBRSOR). We find that existing methods are limited by GPU memory constraints and can only learn local data relationships, and the traditional metric space is suboptimal. To address these limitations, we propose a novel generalized likelihood ratio test-based metric learning (GLRTML) approach that incorporates global statistical information during both the training and testing phases. Our theoretical analyses show that GLRTML estimates the relative difficulty of sample pairs by exploiting global data distribution information, which guides the neural network to focus on difficult samples. Ablation experiments demonstrate the effectiveness of using GLRTML to introduce global information during the training and testing phases.
Accurate estimation of the distribution of embeddings is critical to the effectiveness of GLRTML, and domain shifts between the training and target domains can reduce the effectiveness of using distribution parameters estimated from training data. To overcome this challenge, we propose the clustering pseudo-label-based fast parameter adaptation (CPLFPA) method. CPLFPA efficiently estimates the distribution of embeddings in the target domain by clustering target domain instances and re-estimating the distribution parameters for GLRTML. Experimental results confirm that CPLFPA improves the cross-domain retrieval performance of GLRTML by solving the parameter estimation problem in unsupervised domain adaptation.

\bibliographystyle{IEEEtran}
\bibliography{Ref}

\appendices

\section{Proof of Discussions and Theorem}

\subsection{Proof of Discussion 1}
\renewcommand{\theequation}{A-\arabic{equation}}
\setcounter{equation}{0}

Given false alarm rate $P_{FA} = \rho$, i.e., given hypothesis testing threshold $\beta$:
\begin{equation}
    \hat{s}(i, j) = \log \left( \frac{\hat{p}(\boldsymbol{x}_{\boldsymbol{\theta},ij} \mid \boldsymbol{\vartheta}_1)}{\hat{p}(\boldsymbol{x}_{\boldsymbol{\theta},ij} \mid \boldsymbol{\vartheta}_0)} \right) \gtrless_{\mathcal{H}_0}^{\mathcal{H}_1} \beta.
    \label{eq:A-1}
\end{equation}
MG-GLRTML gives the MLE $\boldsymbol{\vartheta}_1$ and $\boldsymbol{\vartheta}_0$ of distribution parameters in the case of multivariate Gaussian distribution, while GMM-GLRTML gives the local MLE $\boldsymbol{\vartheta}_1$ and $\boldsymbol{\vartheta}_0$ of distribution parameters in the case of Gaussian mixture models modeling distribution. According to the asymptotic nature of MLE~\cite{ref57}:

\begin{equation}
    \lim_{N_1 \to \infty} \mathbb{E}[\boldsymbol{\vartheta}_1] = \boldsymbol{\vartheta}_1^{GT},
    \label{eq:A-2}
\end{equation}

\begin{equation}
    \sqrt{N_1} (\boldsymbol{\vartheta}_1 - \boldsymbol{\vartheta}_1^{GT}) \xrightarrow{d} \mathcal{N}(0, I(\boldsymbol{\vartheta}_1^{GT})^{-1}),
    \label{eq:A-3}
\end{equation}

\begin{equation}
    \lim_{N_1 \to \infty} \mathbb{E}[\boldsymbol{\vartheta}_0] = \boldsymbol{\vartheta}_0^{GT},
    \label{eq:A-4}
\end{equation}

\begin{equation}
    \sqrt{N_0} (\boldsymbol{\vartheta}_0 - \boldsymbol{\vartheta}_0^{GT}) \xrightarrow{d} \mathcal{N}(0, I(\boldsymbol{\vartheta}_0^{GT})^{-1}),
    \label{eq:A-5}
\end{equation}
where $\boldsymbol{\vartheta}_1^{GT}$ and $\boldsymbol{\vartheta}_0^{GT}$ mean the ground truth of distribution parameters, $I(\cdot)$ is the corresponding Fisher information matrix, and $\xrightarrow{d}$ means convergence by distribution. Under the premise that the modeling assumptions of multivariate Gaussian and Gaussian mixture models are accurate, since MLE gives asymptotically effective distribution parameter estimates, the training set and test set are identically distributed, $\hat{p}(\boldsymbol{x}_{\boldsymbol{\theta},ij} \mid \boldsymbol{\vartheta}_1)$ and $\hat{p}(\boldsymbol{x}_{\boldsymbol{\theta},ij} \mid \boldsymbol{\vartheta}_0)$ are accurate estimates for the training set. We define that the arbitrary decision function:

\begin{equation}
    \mathcal{F}(\boldsymbol{x}_{\boldsymbol{\theta},ij}) \triangleq 
    \begin{cases}
        \mathcal{H}_1, & \boldsymbol{x}_{\boldsymbol{\theta},ij} \in R_1 \\
        \mathcal{H}_0, & \boldsymbol{x}_{\boldsymbol{\theta},ij} \in R_0,
    \end{cases}
    \label{eq:A-6}
\end{equation}
where $R_1$ and $R_0$ are decision intervals for hypothesis $\mathcal{H}_1$ and hypothesis $\mathcal{H}_0$, respectively. From the perspective of statistical inference, the optimal retrieval result under a given false alarm rate $P_{FA} = \rho$ is to maximize the probability of detecting relevant targets from the gallery set $P_D$. According to Neyman-Pearson theorem~\cite{ref56}:

\begin{equation}
    \arg\max_F P_D \quad \text{s.t.} \quad P_{FA} = \rho
    \label{eq:A-7}
\end{equation}

\begin{equation}
    \Rightarrow \arg\max_F F = P_D + \lambda (P_{FA} - \rho)
    \label{eq:A-8}
\end{equation}

\begin{equation}
    \Rightarrow \mathcal{F}(\boldsymbol{x}_{\boldsymbol{\theta},ij}) = 
    \begin{cases}
        \mathcal{H}_1, & p(\boldsymbol{x}_{\boldsymbol{\theta},ij} \mid \mathcal{H}_1) + \lambda p(\boldsymbol{x}_{\boldsymbol{\theta},ij} \mid \mathcal{H}_0) > 0 \\
        \mathcal{H}_0, & p(\boldsymbol{x}_{\boldsymbol{\theta},ij} \mid \mathcal{H}_1) + \lambda p(\boldsymbol{x}_{\boldsymbol{\theta},ij} \mid \mathcal{H}_0) < 0
    \end{cases}
    \label{eq:A-9}
\end{equation}
We define $\hat{F} \triangleq \int_{R_1} \left( \hat{p}(\boldsymbol{x}_{\boldsymbol{\theta},ij} \mid \boldsymbol{\vartheta}_1) + \lambda \hat{p}(\boldsymbol{x}_{\boldsymbol{\theta},ij} \mid \boldsymbol{\vartheta}_0) \right) d\boldsymbol{x}_{\boldsymbol{\theta},ij} - \lambda \rho$, considering that $\hat{F} \xrightarrow{d} F$, we have:

\begin{equation}
    \mathcal{F}(\boldsymbol{x}_{\boldsymbol{\theta},ij}) \xrightarrow{d}
    \begin{cases}
        \mathcal{H}_1, & \hat{p}(\boldsymbol{x}_{\boldsymbol{\theta},ij} \mid \boldsymbol{\vartheta}_1) + \lambda \hat{p}(\boldsymbol{x}_{\boldsymbol{\theta},ij} \mid \boldsymbol{\vartheta}_0) > 0 \\
        \mathcal{H}_0, & \hat{p}(\boldsymbol{x}_{\boldsymbol{\theta},ij} \mid \boldsymbol{\vartheta}_1) + \lambda \hat{p}(\boldsymbol{x}_{\boldsymbol{\theta},ij} \mid \boldsymbol{\vartheta}_0) < 0
    \end{cases}
    \label{eq:A-10}
\end{equation}

Equation~\eqref{eq:A-10} shows the asymptotic optimality of Equations~(11) and (15). We use the loss function Equation~(12) to iteratively update the neural network and metric space so that the parameters of the neural network and the metric space are optimized together.

\subsection{Proof of Discussion 2}
\renewcommand{\theequation}{B-\arabic{equation}}
\setcounter{equation}{0}

Consider a batch with $M_1$ and $M_0$ positive samples $\boldsymbol{x}_{\boldsymbol{\theta},m}^{+} \in X_1$ and negative samples $\boldsymbol{x}_{\boldsymbol{\theta},l} \in X_0$, including difficult positive samples $\boldsymbol{x}_{\boldsymbol{\theta},m^*}^{+} \in X_1^*$ and difficult negative samples $\boldsymbol{x}_{\boldsymbol{\theta},l^*} \in X_0^*$, where $m, l, m^*, l^*$ are index variables. Considering that $\hat{p}(\cdot \mid \boldsymbol{\vartheta}_1)$ and $\hat{p}(\cdot \mid \boldsymbol{\vartheta}_0)$ predict difficulty or rarity of embeddings under hypothesis $\mathcal{H}_1$ and hypothesis $\mathcal{H}_0$, we have:

\begin{align}
    \hat{p}(\boldsymbol{x}_{\boldsymbol{\theta},m^*}^+ \mid \boldsymbol{\vartheta}_1) &\ll 
    \hat{p}(\boldsymbol{x}_{\boldsymbol{\theta},m}^+ \mid \boldsymbol{\vartheta}_1), \quad 
    \boldsymbol{x}_{\boldsymbol{\theta},m^*}^+ \in X_1^*, \notag\\
    &\boldsymbol{x}_{\boldsymbol{\theta},m}^+ \in X_1, \, 
    \boldsymbol{x}_{\boldsymbol{\theta},m}^+ \notin X_1^*. 
    \label{eq:B-1}
\end{align}

\begin{align}
    \hat{p}(\boldsymbol{x}_{\boldsymbol{\theta},l^*}^- \mid \boldsymbol{\vartheta}_0) &\ll 
    \hat{p}(\boldsymbol{x}_{\boldsymbol{\theta},l}^- \mid \boldsymbol{\vartheta}_0), \quad 
    \boldsymbol{x}_{\boldsymbol{\theta},l^*}^- \in X_0^*, \notag\\
    &\boldsymbol{x}_{\boldsymbol{\theta},l}^- \in X_0, \, 
    \boldsymbol{x}_{\boldsymbol{\theta},l}^- \notin X_0^*. 
    \label{eq:B-2}
\end{align}
Therefore, let $\hat{s}_p^m$ represent the estimated similarity score of $\boldsymbol{x}_{\boldsymbol{\theta},m}^+ \in X_1, \, \boldsymbol{x}^+_{\boldsymbol{\theta},m} \notin X_1^*$, and $\hat{s}_p^{m^*}$ represent the estimated similarity score of difficult positive samples, we have:

\begin{equation}
    \hat{s}_p^{m^*} < \hat{s}_p^m. 
    \label{eq:B-3}
\end{equation}
In the same way:

\begin{equation}
    \hat{s}_n^{l^*} < \hat{s}_n^l. 
    \label{eq:B-4}
\end{equation}
Let 
\begin{equation}
    h(\hat{s}_n^l, \hat{s}_p^m) = 1 + \sum_{m=0}^{M_1-1}\sum_{l=0}^{M_0-1}
    \exp(\nu(\hat{s}_n^l - \hat{s}_p^m)),
    \label{eq:B-5}
\end{equation}
the partial derivatives of $\mathcal{L}_{\text{GLRTML}}$ with respect to $\hat{s}_p^m$ and $\hat{s}_n^l$ are as follows:

\begin{align}
    \frac{\partial \mathcal{L}_{\text{GLRTML}}}{\partial \hat{s}_p^m} &= 
    \frac{1}{h(\hat{s}_n^l, \hat{s}_p^m)} 
    \frac{\partial h(\hat{s}_n^l, \hat{s}_p^m)}{\partial \hat{s}_p^m} \notag\\
    &= -\nu \exp\left(\nu (-\hat{s}_p^m)\right) \sum_{l=0}^{M_0-1}\exp\left(\nu \hat{s}_n^l\right) \notag\\
    & \quad \times \left(1 + \sum_{m=0}^{M_1-1} \sum_{l=0}^{M_0-1} \exp\left(\nu(\hat{s}_n^l - \hat{s}_p^m)\right)\right)^{-1},
    \label{eq:B-6}
\end{align}

\begin{align}
    \frac{\partial \mathcal{L}_{\text{GLRTML}}}{\partial \hat{s}_n^l} &= 
    \frac{1}{h(\hat{s}_n^l, \hat{s}_p^m)} 
    \frac{\partial h(\hat{s}_n^l, \hat{s}_p^m)}{\partial \hat{s}_n^l} \notag\\
    &= \nu \exp\left(\nu \hat{s}_n^l\right) \sum_{m=0}^{M_1-1} \exp\left(\nu (-\hat{s}_p^m)\right) \notag\\
    & \quad \times \left(1 + \sum_{m=0}^{M_1-1} \sum_{l=0}^{M_0-1} \exp\left(\nu(\hat{s}_n^l - \hat{s}_p^m)\right)\right)^{-1}.
    \label{eq:B-7}
\end{align}

Due to the scaling effect of the exponential function:

\begin{align}
    \left| \frac{\partial \mathcal{L}_{\text{GLRTML}}}{\partial \hat{s}_p^{m^*}} \right| &\gg 
    \left| \frac{\partial \mathcal{L}_{\text{GLRTML}}}{\partial \hat{s}_p^m} \right|, \notag\\
    \left| \frac{\partial \mathcal{L}_{\text{GLRTML}}}{\partial \hat{s}_n^{l^*}} \right| &\gg 
    \left| \frac{\partial \mathcal{L}_{\text{GLRTML}}}{\partial \hat{s}_n^l} \right|.
    \label{eq:B-8}
\end{align}

Therefore, the training process of GLRTML will pay more attention to rare and difficult samples to avoid overfitting.

\subsection{Proof of Theorem 1}
\renewcommand{\theequation}{C-\arabic{equation}}
\setcounter{equation}{0}
For MG-GLRTML, considering that $\hat{\Sigma}_0^{-1} - \hat{\Sigma}_1^{-1}$ is a positive definite matrix, let $\hat{\Sigma}^{-1} = \hat{\Sigma}_0^{-1} - \hat{\Sigma}_1^{-1}$ and simply rewrite Eq.~(11):

\begin{equation}
    \hat{s}(i, j) = \boldsymbol{x}_{\boldsymbol{\theta},ij}^{\top} \hat{\Sigma}^{-1} \boldsymbol{x}_{\boldsymbol{\theta},ij} 
    = D_M^2 \left( \boldsymbol{x}_{\boldsymbol{\theta},ij}, \boldsymbol{0}, \hat{\Sigma} \right).
    \label{eq:C-1}
\end{equation}

For GMM-GLRTML, considering the characteristics of the center distribution of embeddings under hypothesis $\mathcal{H}_1$, we simplify $\hat{p}(\boldsymbol{x}_{\boldsymbol{\theta},ij} \mid \boldsymbol{\vartheta}_1)$ to a zero-mean multivariate Gaussian distribution without losing rationality, rewrite Eq.~(15):

\begin{align}
    \hat{s}(i, j) &= \log \left( 
    \frac{\mathcal{N}(\boldsymbol{x}_{\boldsymbol{\theta},ij} \mid \boldsymbol{0}, \Sigma^1)}
    {\sum_{k=0}^{K_0-1} \pi_k^{0} \mathcal{N}(\boldsymbol{x}_{\boldsymbol{\theta},ij} \mid \boldsymbol{\mu}_k, \Sigma_k^{0})} 
    \right) \notag \\
    &= - \log \left( \sum_{k=0}^{K_0-1} \pi_k^{0} \mathcal{N}(\boldsymbol{x}_{\boldsymbol{\theta},ij} \mid \boldsymbol{\mu}_k, \Sigma_k^{0}) \right) \notag \\
    &\quad - \frac{1}{2} \boldsymbol{x}_{\boldsymbol{\theta},ij}^{\top} (\Sigma^1)^{-1} \boldsymbol{x}_{\boldsymbol{\theta},ij} 
    - \frac{1}{2} \log \left| \Sigma^1 \right| - \frac{d}{2} \log 2\pi.
    \label{eq:C-2}
\end{align}

Suppose the probability that each embedding $\boldsymbol{x}_{\boldsymbol{\theta},ij}$ belongs to each component of hypothesis $\mathcal{H}_0$ is determined by the posterior probability $\gamma_{ij,k}^{0}$. This means that the contribution of each data point to the overall likelihood function can be viewed as the sum of the weighted probabilities of its components. We can use the expectation of log-likelihood to approximate Eq.~\ref{eq:C-2}:

\begin{align}
    \hat{s}(i, j) &= -\sum_{k=0}^{K_0-1} \gamma_{ij,k}^{0} 
    \log \left( \pi_k^{0} \mathcal{N}(\boldsymbol{x}_{\boldsymbol{\theta},ij} \mid \boldsymbol{\mu}_k, \Sigma_k^{0}) \right) \notag \\
    &\quad - \frac{1}{2} \boldsymbol{x}_{\boldsymbol{\theta},ij}^{\top} (\Sigma^1)^{-1} \boldsymbol{x}_{\boldsymbol{\theta},ij} 
    - \frac{1}{2} \log \left| \Sigma^1 \right| - \frac{d}{2} \log 2\pi \notag \\
    &= -\frac{1}{2} \boldsymbol{x}_{\boldsymbol{\theta},ij}^{\top} (\Sigma^1)^{-1} \boldsymbol{x}_{\boldsymbol{\theta},ij} 
    + \sum_{k=0}^{K_0-1} \frac{\gamma_{ij,k}^{0}}{2} 
    (\boldsymbol{x}_{\boldsymbol{\theta},ij} - \boldsymbol{\mu}_k)^{\top} (\Sigma_k^{0})^{-1} \notag \\
    &\quad \times (\boldsymbol{x}_{\boldsymbol{\theta},ij} - \boldsymbol{\mu}_k) + C_1 \notag \\
    &= -\frac{1}{2} D_M^2(\boldsymbol{x}_{\boldsymbol{\theta},ij}, \boldsymbol{0}, \Sigma^1) 
    + \sum_{k=0}^{K_0-1} \frac{\gamma_{ij,k}^{0}}{2} 
    D_M^2 (\boldsymbol{x}_{\boldsymbol{\theta},ij}, \boldsymbol{\mu}_k, \Sigma_k^{0}) + C_1.
    \label{eq:C-3}
\end{align}

Where $C_1$ is a statistically irrelevant part.

\section{Visual Results of different Methods}
\begin{figure*}[!t]
\centering
\includegraphics[width=6.8in]{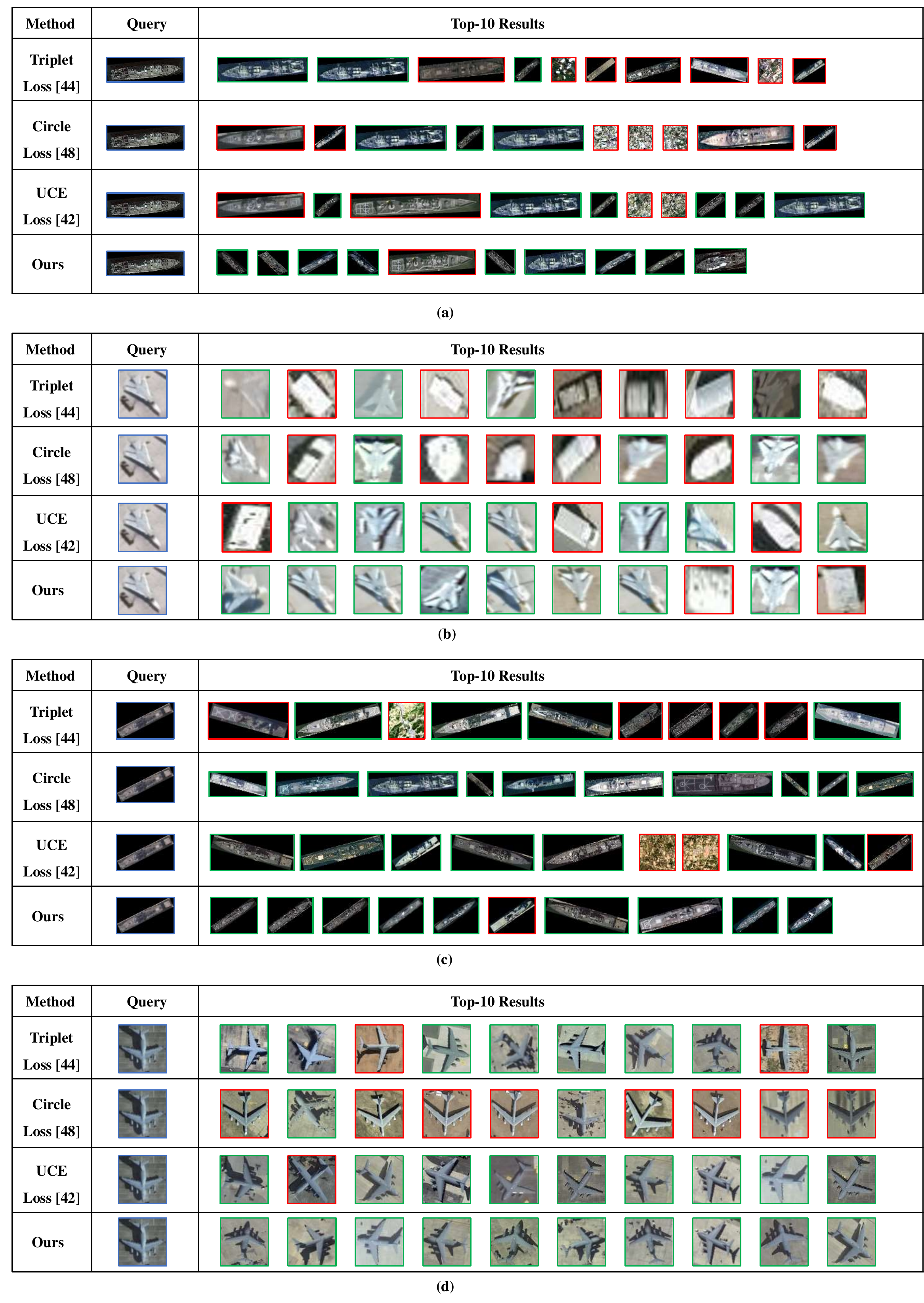}
\caption{Top-10 retrieval results of the proposed method and representative methods on different datasets. (a) Top-10 retrieval results on CBRSOR-FGSRSI. (b) Top-10 retrieval results on CBRSOR-MAR. (c) Top-10 retrieval results on UDA-CBRSOR-FGSRSI. (d) Top-10 retrieval results on UDA-CBRSOR-MAR.}
\label{fig_6}
\end{figure*}

In this section, we present visual comparisons of the Top-10 retrieval results obtained by our proposed method and several representative methods on different datasets, as shown in Fig.~\ref{fig_6}. 

\vfill

\end{document}